\documentclass{article}

    \PassOptionsToPackage{numbers, compress}{natbib}

    \usepackage[preprint]{neurips_2021}

\usepackage[utf8]{inputenc} %
\usepackage[T1]{fontenc}    %
\usepackage{hyperref}       %
\usepackage{url}            %
\usepackage{booktabs}       %
\usepackage{amsfonts}       %
\usepackage{nicefrac}       %
\usepackage{microtype}      %
\usepackage{xcolor}         %

\usepackage[titletoc]{appendix}
\usepackage{placeins}
\usepackage{float}
\usepackage{listings}
\usepackage{bm}
\usepackage{bbm}
\usepackage{amsmath}
\usepackage{amssymb}
\usepackage{amsthm}
\usepackage{enumitem}
\usepackage{graphicx}
\usepackage{scalerel}
\usepackage{mathtools}
\usepackage{tabularx}
\usepackage{tablefootnote}
\usepackage{arydshln,leftidx}
\usepackage{wrapfig}
\usepackage{multirow}
\usepackage{ragged2e}  %
\newcolumntype{Y}{>{\RaggedRight\arraybackslash}X}

\usepackage{color}
\definecolor{myfavblue}{rgb}{0.05, 0.2, 0.8}
\definecolor{keywords}{RGB}{255,0,90}
\definecolor{comments}{RGB}{0,0,113}
\definecolor{red}{RGB}{160,0,0}
\definecolor{green}{RGB}{0,150,0}
\definecolor{C0}{rgb}{0.12156862745098039, 0.4666666666666667, 0.7058823529411765}  %

\definecolor{myblue}{HTML}{3182bd}
\definecolor{myred}{HTML}{de2d26}

\usepackage{hyperref}
\usepackage{url}
\usepackage{relsize}
\usepackage{verbatim}
\usepackage{tablefootnote}

\usepackage[frozencache]{minted}

\usepackage{tikz}
\usetikzlibrary{fit,positioning}
\usetikzlibrary{calc,arrows.meta,positioning, arrows}
\usetikzlibrary{shapes.geometric}

\tikzset{
    every node/.style={font=\sffamily\Large},
    main node/.style={thick,circle,draw,font=\sffamily\huge}
}

\definecolor{mydarkblue}{rgb}{0,0.08,0.45}
\hypersetup{
    colorlinks=true,
    linkcolor=mydarkblue,
    citecolor=mydarkblue,
    filecolor=mydarkblue,
    urlcolor=mydarkblue
}

\usepackage{listings}
\usepackage{tikz}
\usetikzlibrary{arrows.meta}

\newcounter{tmlistings}

\newcommand\makenode[2]{%
  \tikz[baseline=0pt, remember picture] { \node[fill=gray!50,thick,rounded corners,anchor=base,#1/.try] (listings-\the\value{tmlistings}) {#2}; }%
  \stepcounter{tmlistings}%
}

\tikzset{
  keyword/.style={
    fill=gray!75,
    draw=black
  }
}

\usepackage{caption}
\usepackage{xargs}                      %
\usepackage[colorinlistoftodos,prependcaption,textsize=tiny,textwidth=15mm]{todonotes}
\newcommandx{\unsure}[2][1=]{\todo[linecolor=red,backgroundcolor=red!25,bordercolor=red,#1]{#2}}
\newcommandx{\change}[2][1=]{\todo[linecolor=blue,backgroundcolor=blue!25,bordercolor=blue,#1]{#2}}
\newcommandx{\info}[2][1=]{\todo[linecolor=OliveGreen,backgroundcolor=OliveGreen!25,bordercolor=OliveGreen,#1]{#2}}
\newcommandx{\improvement}[2][1=]{\todo[linecolor=Plum,backgroundcolor=Plum!25,bordercolor=Plum,#1]{#2}}
\newcommandx{\thiswillnotshow}[2][1=]{\todo[disable,#1]{#2}}
\reversemarginpar

\title{
\Large
Learning to Extend Program Graphs to Work-in-Progress Code
}

\author{%
  Xuechen Li\thanks{Work done during Google AI Residency.} \\
  Stanford University\\
  \texttt{lxuechen@stanford.edu} \\
   \And
   Chris J. Maddison \\
   University of Toronto \& Vector Institute \\
   \texttt{cmaddis@cs.toronto.edu} \\
   \AND
   Daniel Tarlow \\
   Google Research \\
   \texttt{dtarlow@google.com} \\
}

\begin{document}

\maketitle

\newcommand{\figleft}{{\em (Left)}}
\newcommand{\figcenter}{{\em (Center)}}
\newcommand{\figright}{{\em (Right)}}
\newcommand{\figtop}{{\em (Top)}}
\newcommand{\figbottom}{{\em (Bottom)}}
\newcommand{\captiona}{{\em (a)}}
\newcommand{\captionb}{{\em (b)}}
\newcommand{\captionc}{{\em (c)}}
\newcommand{\captiond}{{\em (d)}}

\newcommand{\newterm}[1]{{\bf #1}}

\def\figref#1{figure~\ref{#1}}
\def\Figref#1{Figure~\ref{#1}}
\def\twofigref#1#2{figures \ref{#1} and \ref{#2}}
\def\quadfigref#1#2#3#4{figures \ref{#1}, \ref{#2}, \ref{#3} and \ref{#4}}
\def\secref#1{section~\ref{#1}}
\def\Secref#1{Section~\ref{#1}}
\def\twosecrefs#1#2{sections \ref{#1} and \ref{#2}}
\def\secrefs#1#2#3{sections \ref{#1}, \ref{#2} and \ref{#3}}
\def\eqref#1{equation~\ref{#1}}
\def\Eqref#1{Equation~\ref{#1}}
\def\plaineqref#1{\ref{#1}}
\def\chapref#1{chapter~\ref{#1}}
\def\Chapref#1{Chapter~\ref{#1}}
\def\rangechapref#1#2{chapters\ref{#1}--\ref{#2}}
\def\partref#1{part~\ref{#1}}
\def\Partref#1{Part~\ref{#1}}
\def\twopartref#1#2{parts \ref{#1} and \ref{#2}}

\def\ceil#1{\lceil #1 \rceil}
\def\floor#1{\lfloor #1 \rfloor}
\def\1{\bm{1}}
\newcommand{\train}{\mathcal{D}}
\newcommand{\valid}{\mathcal{D_{\mathrm{valid}}}}
\newcommand{\test}{\mathcal{D_{\mathrm{test}}}}

\def\eps{{\epsilon}}

\def\reta{{\textnormal{$\eta$}}}
\def\ra{{\textnormal{a}}}
\def\rb{{\textnormal{b}}}
\def\rc{{\textnormal{c}}}
\def\rd{{\textnormal{d}}}
\def\re{{\textnormal{e}}}
\def\rf{{\textnormal{f}}}
\def\rg{{\textnormal{g}}}
\def\rh{{\textnormal{h}}}
\def\ri{{\textnormal{i}}}
\def\rj{{\textnormal{j}}}
\def\rk{{\textnormal{k}}}
\def\rl{{\textnormal{l}}}
\def\rn{{\textnormal{n}}}
\def\ro{{\textnormal{o}}}
\def\rp{{\textnormal{p}}}
\def\rq{{\textnormal{q}}}
\def\rr{{\textnormal{r}}}
\def\rs{{\textnormal{s}}}
\def\rt{{\textnormal{t}}}
\def\ru{{\textnormal{u}}}
\def\rv{{\textnormal{v}}}
\def\rw{{\textnormal{w}}}
\def\rx{{\textnormal{x}}}
\def\ry{{\textnormal{y}}}
\def\rz{{\textnormal{z}}}

\def\rvepsilon{{\mathbf{\epsilon}}}
\def\rvtheta{{\mathbf{\theta}}}
\def\rva{{\mathbf{a}}}
\def\rvb{{\mathbf{b}}}
\def\rvc{{\mathbf{c}}}
\def\rvd{{\mathbf{d}}}
\def\rve{{\mathbf{e}}}
\def\rvf{{\mathbf{f}}}
\def\rvg{{\mathbf{g}}}
\def\rvh{{\mathbf{h}}}
\def\rvu{{\mathbf{i}}}
\def\rvj{{\mathbf{j}}}
\def\rvk{{\mathbf{k}}}
\def\rvl{{\mathbf{l}}}
\def\rvm{{\mathbf{m}}}
\def\rvn{{\mathbf{n}}}
\def\rvo{{\mathbf{o}}}
\def\rvp{{\mathbf{p}}}
\def\rvq{{\mathbf{q}}}
\def\rvr{{\mathbf{r}}}
\def\rvs{{\mathbf{s}}}
\def\rvt{{\mathbf{t}}}
\def\rvu{{\mathbf{u}}}
\def\rvv{{\mathbf{v}}}
\def\rvw{{\mathbf{w}}}
\def\rvx{{\mathbf{x}}}
\def\rvy{{\mathbf{y}}}
\def\rvz{{\mathbf{z}}}

\def\erva{{\textnormal{a}}}
\def\ervb{{\textnormal{b}}}
\def\ervc{{\textnormal{c}}}
\def\ervd{{\textnormal{d}}}
\def\erve{{\textnormal{e}}}
\def\ervf{{\textnormal{f}}}
\def\ervg{{\textnormal{g}}}
\def\ervh{{\textnormal{h}}}
\def\ervi{{\textnormal{i}}}
\def\ervj{{\textnormal{j}}}
\def\ervk{{\textnormal{k}}}
\def\ervl{{\textnormal{l}}}
\def\ervm{{\textnormal{m}}}
\def\ervn{{\textnormal{n}}}
\def\ervo{{\textnormal{o}}}
\def\ervp{{\textnormal{p}}}
\def\ervq{{\textnormal{q}}}
\def\ervr{{\textnormal{r}}}
\def\ervs{{\textnormal{s}}}
\def\ervt{{\textnormal{t}}}
\def\ervu{{\textnormal{u}}}
\def\ervv{{\textnormal{v}}}
\def\ervw{{\textnormal{w}}}
\def\ervx{{\textnormal{x}}}
\def\ervy{{\textnormal{y}}}
\def\ervz{{\textnormal{z}}}

\def\rmA{{\mathbf{A}}}
\def\rmB{{\mathbf{B}}}
\def\rmC{{\mathbf{C}}}
\def\rmD{{\mathbf{D}}}
\def\rmE{{\mathbf{E}}}
\def\rmF{{\mathbf{F}}}
\def\rmG{{\mathbf{G}}}
\def\rmH{{\mathbf{H}}}
\def\rmI{{\mathbf{I}}}
\def\rmJ{{\mathbf{J}}}
\def\rmK{{\mathbf{K}}}
\def\rmL{{\mathbf{L}}}
\def\rmM{{\mathbf{M}}}
\def\rmN{{\mathbf{N}}}
\def\rmO{{\mathbf{O}}}
\def\rmP{{\mathbf{P}}}
\def\rmQ{{\mathbf{Q}}}
\def\rmR{{\mathbf{R}}}
\def\rmS{{\mathbf{S}}}
\def\rmT{{\mathbf{T}}}
\def\rmU{{\mathbf{U}}}
\def\rmV{{\mathbf{V}}}
\def\rmW{{\mathbf{W}}}
\def\rmX{{\mathbf{X}}}
\def\rmY{{\mathbf{Y}}}
\def\rmZ{{\mathbf{Z}}}

\def\ermA{{\textnormal{A}}}
\def\ermB{{\textnormal{B}}}
\def\ermC{{\textnormal{C}}}
\def\ermD{{\textnormal{D}}}
\def\ermE{{\textnormal{E}}}
\def\ermF{{\textnormal{F}}}
\def\ermG{{\textnormal{G}}}
\def\ermH{{\textnormal{H}}}
\def\ermI{{\textnormal{I}}}
\def\ermJ{{\textnormal{J}}}
\def\ermK{{\textnormal{K}}}
\def\ermL{{\textnormal{L}}}
\def\ermM{{\textnormal{M}}}
\def\ermN{{\textnormal{N}}}
\def\ermO{{\textnormal{O}}}
\def\ermP{{\textnormal{P}}}
\def\ermQ{{\textnormal{Q}}}
\def\ermR{{\textnormal{R}}}
\def\ermS{{\textnormal{S}}}
\def\ermT{{\textnormal{T}}}
\def\ermU{{\textnormal{U}}}
\def\ermV{{\textnormal{V}}}
\def\ermW{{\textnormal{W}}}
\def\ermX{{\textnormal{X}}}
\def\ermY{{\textnormal{Y}}}
\def\ermZ{{\textnormal{Z}}}

\def\vzero{{\bm{0}}}
\def\vone{{\bm{1}}}
\def\vmu{{\bm{\mu}}}
\def\vtheta{{\bm{\theta}}}
\def\va{{\bm{a}}}
\def\vb{{\bm{b}}}
\def\vc{{\bm{c}}}
\def\vd{{\bm{d}}}
\def\ve{{\bm{e}}}
\def\vf{{\bm{f}}}
\def\vg{{\bm{g}}}
\def\vh{{\bm{h}}}
\def\vi{{\bm{i}}}
\def\vj{{\bm{j}}}
\def\vk{{\bm{k}}}
\def\vl{{\bm{l}}}
\def\vm{{\bm{m}}}
\def\vn{{\bm{n}}}
\def\vo{{\bm{o}}}
\def\vp{{\bm{p}}}
\def\vq{{\bm{q}}}
\def\vr{{\bm{r}}}
\def\vs{{\bm{s}}}
\def\vt{{\bm{t}}}
\def\vu{{\bm{u}}}
\def\vv{{\bm{v}}}
\def\vw{{\bm{w}}}
\def\vx{{\bm{x}}}
\def\vy{{\bm{y}}}
\def\vz{{\bm{z}}}

\def\evalpha{{\alpha}}
\def\evbeta{{\beta}}
\def\evepsilon{{\epsilon}}
\def\evlambda{{\lambda}}
\def\evomega{{\omega}}
\def\evmu{{\mu}}
\def\evpsi{{\psi}}
\def\evsigma{{\sigma}}
\def\evtheta{{\theta}}
\def\eva{{a}}
\def\evb{{b}}
\def\evc{{c}}
\def\evd{{d}}
\def\eve{{e}}
\def\evf{{f}}
\def\evg{{g}}
\def\evh{{h}}
\def\evi{{i}}
\def\evj{{j}}
\def\evk{{k}}
\def\evl{{l}}
\def\evm{{m}}
\def\evn{{n}}
\def\evo{{o}}
\def\evp{{p}}
\def\evq{{q}}
\def\evr{{r}}
\def\evs{{s}}
\def\evt{{t}}
\def\evu{{u}}
\def\evv{{v}}
\def\evw{{w}}
\def\evx{{x}}
\def\evy{{y}}
\def\evz{{z}}

\def\mA{{\bm{A}}}
\def\mB{{\bm{B}}}
\def\mC{{\bm{C}}}
\def\mD{{\bm{D}}}
\def\mE{{\bm{E}}}
\def\mF{{\bm{F}}}
\def\mG{{\bm{G}}}
\def\mH{{\bm{H}}}
\def\mI{{\bm{I}}}
\def\mJ{{\bm{J}}}
\def\mK{{\bm{K}}}
\def\mL{{\bm{L}}}
\def\mM{{\bm{M}}}
\def\mN{{\bm{N}}}
\def\mO{{\bm{O}}}
\def\mP{{\bm{P}}}
\def\mQ{{\bm{Q}}}
\def\mR{{\bm{R}}}
\def\mS{{\bm{S}}}
\def\mT{{\bm{T}}}
\def\mU{{\bm{U}}}
\def\mV{{\bm{V}}}
\def\mW{{\bm{W}}}
\def\mX{{\bm{X}}}
\def\mY{{\bm{Y}}}
\def\mZ{{\bm{Z}}}
\def\mBeta{{\bm{\beta}}}
\def\mPhi{{\bm{\Phi}}}
\def\mLambda{{\bm{\Lambda}}}
\def\mSigma{{\bm{\Sigma}}}

\newcommand{\tens}[1]{\bm{\mathsfit{#1}}}
\def\tA{{\tens{A}}}
\def\tB{{\tens{B}}}
\def\tC{{\tens{C}}}
\def\tD{{\tens{D}}}
\def\tE{{\tens{E}}}
\def\tF{{\tens{F}}}
\def\tG{{\tens{G}}}
\def\tH{{\tens{H}}}
\def\tI{{\tens{I}}}
\def\tJ{{\tens{J}}}
\def\tK{{\tens{K}}}
\def\tL{{\tens{L}}}
\def\tM{{\tens{M}}}
\def\tN{{\tens{N}}}
\def\tO{{\tens{O}}}
\def\tP{{\tens{P}}}
\def\tQ{{\tens{Q}}}
\def\tR{{\tens{R}}}
\def\tS{{\tens{S}}}
\def\tT{{\tens{T}}}
\def\tU{{\tens{U}}}
\def\tV{{\tens{V}}}
\def\tW{{\tens{W}}}
\def\tX{{\tens{X}}}
\def\tY{{\tens{Y}}}
\def\tZ{{\tens{Z}}}

\def\gA{{\mathcal{A}}}
\def\gB{{\mathcal{B}}}
\def\gC{{\mathcal{C}}}
\def\gD{{\mathcal{D}}}
\def\gE{{\mathcal{E}}}
\def\gF{{\mathcal{F}}}
\def\gG{{\mathcal{G}}}
\def\gH{{\mathcal{H}}}
\def\gI{{\mathcal{I}}}
\def\gJ{{\mathcal{J}}}
\def\gK{{\mathcal{K}}}
\def\gL{{\mathcal{L}}}
\def\gM{{\mathcal{M}}}
\def\gN{{\mathcal{N}}}
\def\gO{{\mathcal{O}}}
\def\gP{{\mathcal{P}}}
\def\gQ{{\mathcal{Q}}}
\def\gR{{\mathcal{R}}}
\def\gS{{\mathcal{S}}}
\def\gT{{\mathcal{T}}}
\def\gU{{\mathcal{U}}}
\def\gV{{\mathcal{V}}}
\def\gW{{\mathcal{W}}}
\def\gX{{\mathcal{X}}}
\def\gY{{\mathcal{Y}}}
\def\gZ{{\mathcal{Z}}}

\def\sA{{\mathbb{A}}}
\def\sB{{\mathbb{B}}}
\def\sC{{\mathbb{C}}}
\def\sD{{\mathbb{D}}}
\def\sF{{\mathbb{F}}}
\def\sG{{\mathbb{G}}}
\def\sH{{\mathbb{H}}}
\def\sI{{\mathbb{I}}}
\def\sJ{{\mathbb{J}}}
\def\sK{{\mathbb{K}}}
\def\sL{{\mathbb{L}}}
\def\sM{{\mathbb{M}}}
\def\sN{{\mathbb{N}}}
\def\sO{{\mathbb{O}}}
\def\sP{{\mathbb{P}}}
\def\sQ{{\mathbb{Q}}}
\def\sR{{\mathbb{R}}}
\def\sS{{\mathbb{S}}}
\def\sT{{\mathbb{T}}}
\def\sU{{\mathbb{U}}}
\def\sV{{\mathbb{V}}}
\def\sW{{\mathbb{W}}}
\def\sX{{\mathbb{X}}}
\def\sY{{\mathbb{Y}}}
\def\sZ{{\mathbb{Z}}}

\def\emLambda{{\Lambda}}
\def\emA{{A}}
\def\emB{{B}}
\def\emC{{C}}
\def\emD{{D}}
\def\emE{{E}}
\def\emF{{F}}
\def\emG{{G}}
\def\emH{{H}}
\def\emI{{I}}
\def\emJ{{J}}
\def\emK{{K}}
\def\emL{{L}}
\def\emM{{M}}
\def\emN{{N}}
\def\emO{{O}}
\def\emP{{P}}
\def\emQ{{Q}}
\def\emR{{R}}
\def\emS{{S}}
\def\emT{{T}}
\def\emU{{U}}
\def\emV{{V}}
\def\emW{{W}}
\def\emX{{X}}
\def\emY{{Y}}
\def\emZ{{Z}}
\def\emSigma{{\Sigma}}

\newcommand{\etens}[1]{\mathsfit{#1}}
\def\etLambda{{\etens{\Lambda}}}
\def\etA{{\etens{A}}}
\def\etB{{\etens{B}}}
\def\etC{{\etens{C}}}
\def\etD{{\etens{D}}}
\def\etE{{\etens{E}}}
\def\etF{{\etens{F}}}
\def\etG{{\etens{G}}}
\def\etH{{\etens{H}}}
\def\etI{{\etens{I}}}
\def\etJ{{\etens{J}}}
\def\etK{{\etens{K}}}
\def\etL{{\etens{L}}}
\def\etM{{\etens{M}}}
\def\etN{{\etens{N}}}
\def\etO{{\etens{O}}}
\def\etP{{\etens{P}}}
\def\etQ{{\etens{Q}}}
\def\etR{{\etens{R}}}
\def\etS{{\etens{S}}}
\def\etT{{\etens{T}}}
\def\etU{{\etens{U}}}
\def\etV{{\etens{V}}}
\def\etW{{\etens{W}}}
\def\etX{{\etens{X}}}
\def\etY{{\etens{Y}}}
\def\etZ{{\etens{Z}}}

\newcommand{\pdata}{p_{\rm{data}}}
\newcommand{\ptrain}{\hat{p}_{\rm{data}}}
\newcommand{\Ptrain}{\hat{P}_{\rm{data}}}
\newcommand{\pmodel}{p_{\rm{model}}}
\newcommand{\Pmodel}{P_{\rm{model}}}
\newcommand{\ptildemodel}{\tilde{p}_{\rm{model}}}
\newcommand{\pencode}{p_{\rm{encoder}}}
\newcommand{\pdecode}{p_{\rm{decoder}}}
\newcommand{\precons}{p_{\rm{reconstruct}}}

\newcommand{\laplace}{\mathrm{Laplace}} %

\newcommand{\E}{\mathbb{E}}
\newcommand{\Ls}{\mathcal{L}}
\newcommand{\R}{\mathbb{R}}
\newcommand{\emp}{\tilde{p}}
\newcommand{\lr}{\alpha}
\newcommand{\reg}{\lambda}
\newcommand{\rect}{\mathrm{rectifier}}
\newcommand{\softmax}{\mathrm{softmax}}
\newcommand{\sigmoid}{\sigma}
\newcommand{\softplus}{\zeta}
\newcommand{\KL}{D_{\mathrm{KL}}}
\newcommand{\Var}{\mathrm{Var}}
\newcommand{\standarderror}{\mathrm{SE}}
\newcommand{\Cov}{\mathrm{Cov}}
\newcommand{\normlzero}{L^0}
\newcommand{\normlone}{L^1}
\newcommand{\normltwo}{L^2}
\newcommand{\normlp}{L^p}
\newcommand{\normmax}{L^\infty}

\newcommand{\parents}{Pa} %

\let\ab\allowbreak

\newcommand{\eq}[1]{\begin{align}#1\end{align}}
\newcommand{\eqn}[1]{\begin{align*}#1\end{align*}}
\newcommand{\mypmatrix}[1]{\begin{pmatrix}#1\end{pmatrix}}

\newcommand{\dee}{\mathop{\mathrm{d}\!}}
\newcommand{\dt}{\,\dee t}
\newcommand{\de}{\,\dee e}
\newcommand{\ds}{\,\dee s}
\newcommand{\dx}{\,\dee x}
\newcommand{\dX}{\,\dee X}
\newcommand{\dy}{\,\dee y}
\newcommand{\dY}{\,\dee Y}
\newcommand{\dz}{\,\dee z}
\newcommand{\dZ}{\,\dee Z}
\newcommand{\dv}{\,\dee v}
\newcommand{\du}{\,\dee u}
\newcommand{\dw}{\,\dee w}
\newcommand{\dr}{\,\dee r}
\newcommand{\dB}{\,\dee B} %
\newcommand{\db}{\,\dee b}
\newcommand{\dW}{\,\dee W} %
\newcommand{\dtau}{\,\dee \tau}
\newcommand{\dmu}{\,\dee \mu}
\newcommand{\dnu}{\,\dee \nu}
\newcommand{\dzeta}{\,\dee \zeta}
\newcommand{\domega}{\,\dee \omega}

\newcommand*{\matr}[1]{\mathbfit{#1}}
\newcommand*{\tran}{^\top}
\newcommand*{\conj}[1]{\overline{#1}}
\newcommand*{\hermconj}{^{\mathsf{H}}}

\newcommand{\latent}{\vz}
\newcommand{\hidden}{\vh}
\newcommand{\obs}{x}
\newcommand{\sol}{z}
\newcommand{\obsdim}{D_x}
\newcommand{\latentdim}{D}
\newcommand{\solvefunc}{\textnormal{ODESolve}}
\newcommand{\tstart}{{t_\textnormal{0}}}
\newcommand{\tend}{{t_\textnormal{1}}}
\newcommand{\lograte}{\lambda}%
\newcommand{\method}{Latent ODE}
\newcommand{\cnfx}{\sol}
\newcommand{\adj}{a}

\newcommand*{\A}{\mathcal{A}}
\newcommand*{\B}{\mathcal{B}}
\newcommand*{\F}{\mathcal{F}}
\newcommand*{\V}{\mathbb{V}}
\newcommand*{\N}{\mathcal{N}}
\newcommand*{\TV}{\text{TV}}
\newcommand*{\LL}{\left}
\newcommand*{\RR}{\right}
\newcommand*{\tPhi}{\tilde{\Phi}}

\newcommand*{\bracks}[1]{\left(#1\right)}  %
\newcommand*{\abracks}[1]{\left\langle#1\right\rangle}  %
\newcommand*{\sbracks}[1]{\left[#1\right]}  %
\newcommand*{\norm}[1]{\left\|#1 \right\|}  %

\newcommand{\plim}{\mathrm{plim}}

\newtheorem{defi}{Definition}
\numberwithin{defi}{section}
\newtheorem{mycond}{Condition}

\newtheorem{fact}[defi]{Fact}
\newtheorem{cond}[defi]{Condition}
\newtheorem{prop}[defi]{Proposition}

\newtheorem{lemm}[defi]{Lemma}
\newtheorem{theo}[defi]{Theorem}
\newtheorem{coro}[defi]{Corollary}

\newtheorem{remark}{\textbf{Remark}}

\newtheorem{exam}{\textbf{Example}}

\newtheorem{assu}[defi]{Assumption}

\makeatletter
\DeclareRobustCommand\widecheck[1]{{\mathpalette\@widecheck{#1}}}
\def\@widecheck#1#2{%
    \setbox\z@\hbox{\m@th$#1#2$}%
    \setbox\tw@\hbox{\m@th$#1%
       \widehat{%
          \vrule\@width\z@\@height\ht\z@
          \vrule\@height\z@\@width\wd\z@}$}%
    \dp\tw@-\ht\z@
    \@tempdima\ht\z@ \advance\@tempdima2\ht\tw@ \divide\@tempdima\thr@@
    \setbox\tw@\hbox{%
       \raise\@tempdima\hbox{\scalebox{1}[-1]{\lower\@tempdima\box
\tw@}}}%
    {\ooalign{\box\tw@ \cr \box\z@}}}
\makeatother

\def\ssum{\mathsmaller{\sum}}
\def\sint{\mathsmaller{\int}}

\newcommand{\Exp}[1]{ \mathbb{E} \left[ #1 \right] }
\newcommand{\loss}{ \mathcal{L} }
\newcommand{\param}{ {\bm{\theta}} }

\newcommand{\mymid}{ {\text{mid}} }
\DeclarePairedDelimiter\abs{\lvert}{\rvert}

\begin{abstract}

Source code spends most of its time in a broken or incomplete state during software development.
This presents a challenge to machine learning for code, since high-performing models typically rely on graph structured representations of programs derived from traditional program analyses. Such analyses may be undefined for broken or incomplete code.
We extend the notion of program graphs to work-in-progress code by learning to predict edge relations between tokens, training on well-formed code before transferring to work-in-progress code.
We consider the tasks of code completion and localizing and repairing variable misuse in a work-in-process scenario.
We demonstrate that training relation-aware models with fine-tuned edges consistently leads to improved performance on both tasks.

\end{abstract}

\section{Introduction}
Source code written by people is natural in the sense of being predictable~\cite{hindle2012naturalness,ray2016naturalness}. 
This warrants building machine learning models for code~\cite{allamanis2018survey}. These models range from ones based on n-grams~\cite{hindle2012naturalness,hellendoorn2017deep} and graphical models~\cite{raychev2015predicting}, to those based on various neural networks~\cite{allamanis2017learning,brockschmidt2018generative,bieber2020learning,2021languageagnostic,kim2020code,shiv2019novel,zhang2019novel,mou2016convolutional,alon2018code2seq,alon2019code2vec,chen2018tree,gupta2017deepfix} and combinations thereof~\cite{hellendoorn2019global,tarlow2020learning,yasunaga2020graph}. 

While promising results can be obtained by building models of the source code token sequence alone, domain-specific task performance can typically be improved by leveraging graph-structured representations of programs,
where edge relations are derived from traditional program analyses~\cite{allamanis2017learning,ben2018neural,cvitkovic2019open,cummins2020programl}.
Given well-formed code, a variety of program analyses can be run to determine properties of code semantics and relationships between code elements~\cite{cousot1977abstract,reps1998program,nielson2004principles}.
However, for work-in-progress code that may be broken or incomplete, 
these analyses are typically undefined under the programming language's standard specification. 
\begin{wrapfigure}[10]{r}{0.39\textwidth}
\centering
\includegraphics[width=0.35\textwidth]{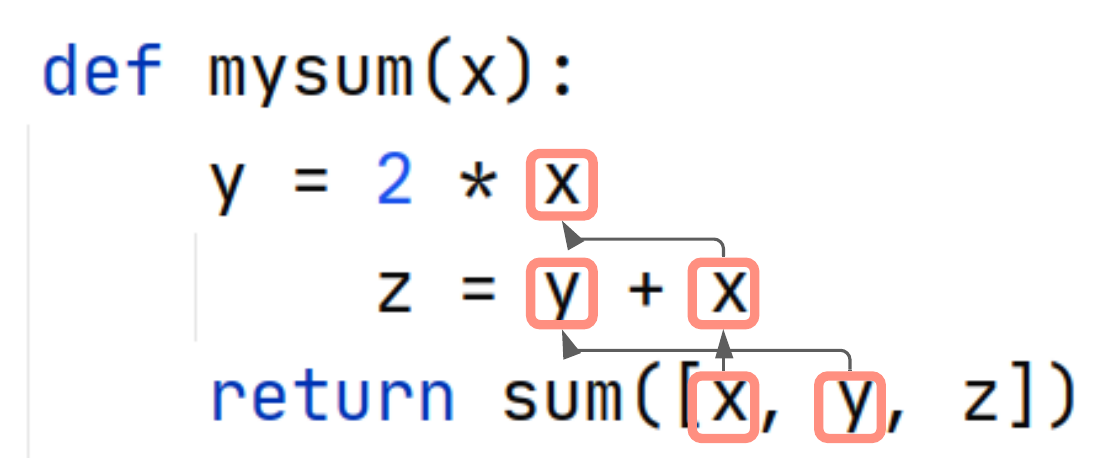}
\caption{
Our relation model predicts LastRead edges sensibly for a piece of code with invalid indentation.
}
\label{fig:fig1}
\end{wrapfigure}

\vspace{-3mm}
One could attempt to build program graph libraries on top of robust analyzers \cite{joern}
or by training a machine learning model to repair code and then analyzing the fixed code.
Yet, both approaches create a new hard subproblem that may not be necessary
to solve perfectly in order to obtain good performance on the downstream task.
Instead, we propose a simpler option that requires only the ability to run program analyses on well-formed code, and results in model components that can be fined-tuned while training a relation-aware model on a downstream task.

Consider the example in Figure~\ref{fig:fig1} where there is an extra indentation in a piece of Python code. This code cannot be executed, and thus it does not strictly make sense to analyze properties of its semantics.
However, one could reasonably expect that the data flow relations illustrated by the arrows still hold.
In fact, these arrows are predictions made by our proposed relation prediction model, trained on well-formed code and then transferred to the broken code snippet in the figure.

Our primary contribution is to show that graph-structured representations of programs derived from traditional program analyses can be extended to broken and incomplete code by leveraging inherent generalization in machine learning models that predict edges. 
We train these models with direct supervision on well-formed code before transferring to work-in-progress code. 
We demonstrate the effectiveness of this out-of-distribution generalization by showing that relation-aware models~\cite{shaw2018self,yang2019xlnet} can effectively use the predicted edges on work-in-process code.
We evaluate our approach on two downstream tasks where work-in-progress code is broken in different ways: \textit{(a)} localizing and repairing variable misuses and \textit{(b)} code completion.
By additionally fine-tuning the relation prediction model during task learning, we achieve strong improvements over relation-agnostic baselines. 
In total, this work points towards a simple and practical solution for realizing the power of relation-aware models on the work-in-progress code that commonly arises in the software development process.

\vspace{-2mm}
\section{Background}
We review Transformers and relation-aware models for code, on which our work is heavily based. 

\vspace{-2mm}
\paragraph{Transformer.}
The Transformer model~\cite{vaswani2017attention} consists of stacked multi-head attention and position-wise fully-connected layers. 
A multi-head attention layer enables encodings at a position in subsequent layers to attend to encodings at all locations in the previous layer via softmax attention whose scores are computed as
$
\bm{\alpha} = \text{Softmax}( \mathbf{q}^\top \mathbf{k} / \sqrt{d} ).
$
Here, $\bm{\alpha}$ is the attention score, $\mathbf{q}$ and $\mathbf{k}$ are respectively the query and key, and $d$ is the dimension of the encoding space. 

\vspace{-2mm}
\paragraph{Relation-aware Models for Source Code.}
Many machine learning models for code take into account relational structure. 
These models include variants of graph neural networks~\cite{allamanis2017learning} and Transformers using relative positions~\cite{shaw2018self,hellendoorn2019global,wang2019rat,2021languageagnostic}. 
Our focus is specifically on models that modify the standard attention computation~\cite{shaw2018self,dai2019transformer} by altering $\mathbf{q}$ or $\mathbf{k}$ using edge embeddings. 
We use the terms \textit{relation-aware} and \textit{edge-aware} exclusively for these models in the paper for ease of presentation. 

\vspace{-2mm}
\section{Learning to Extend Program Graphs}
\vspace{-2mm}
We propose to learn a relation prediction model for source code tokens so that relation-aware models can be applied to work-in-progress code which may be broken or incomplete. 
We cast the learning problem as a supervised one: Given ground truth edges between tokens, train a model to predict edges given only the token sequence. This is a multi-label classification problem, where multiple edges of different types might exist between the same ordered token pair. 

\subsection{Ground Truth Relations}\label{subsec:ground_truth_relations}
\begin{wrapfigure}[13]{R}{0.42\textwidth}
\vspace{-8mm}
\centering
{\includegraphics[width=0.38\textwidth]{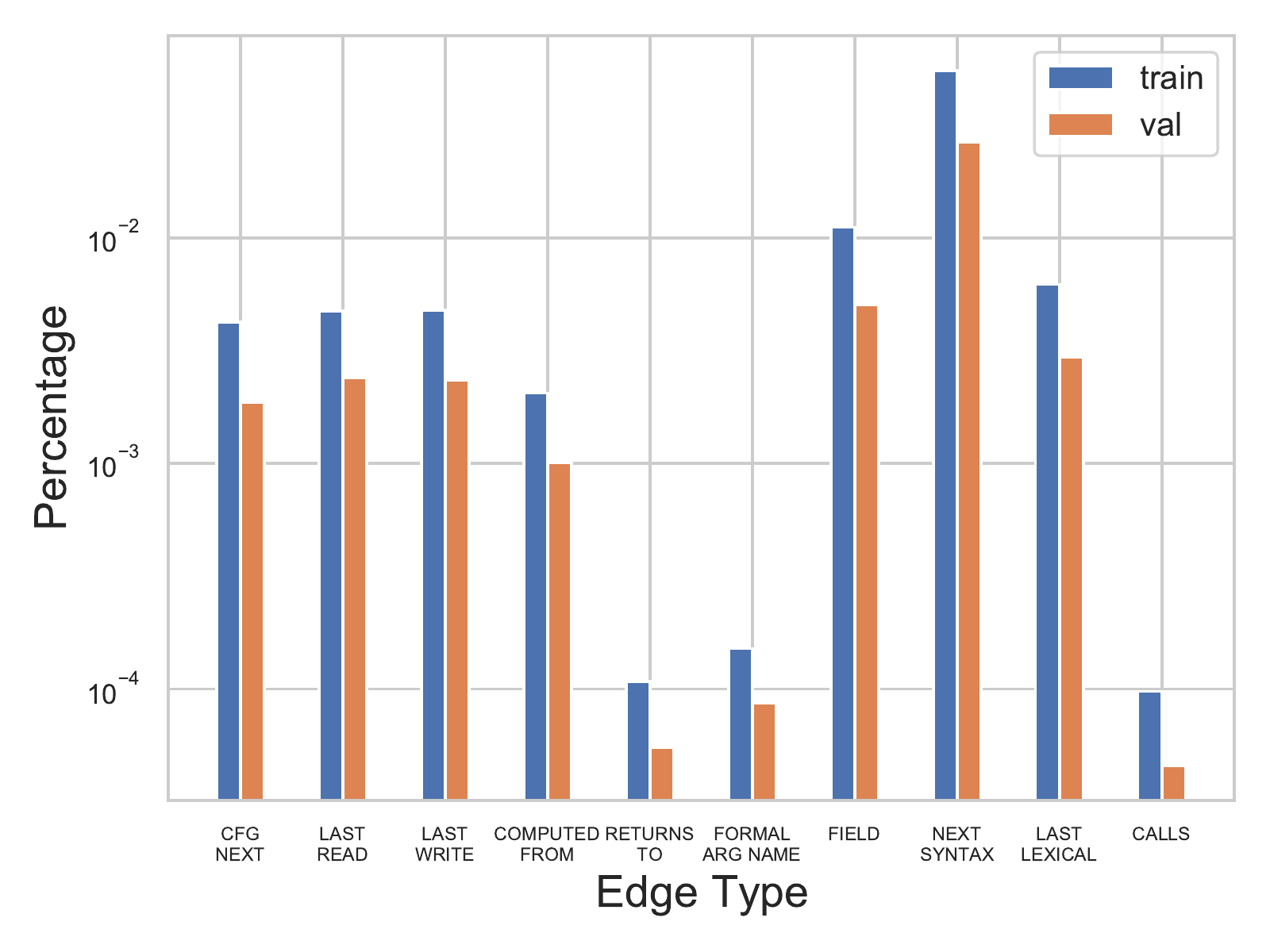}}
\caption{
Edge occurrence frequencies on the large training and validation sets from the Python corpus~\citep{karampatsis2020big}. 
}
\label{fig:edge_on_freq_completion_large}
\end{wrapfigure}
We use the edge relations described by~\citet{hellendoorn2019global} as the ground truth.
The relations are derived from ASTs, control flow graphs, and data flow graphs of well-formed code and are of $10$ types.
Not all types have similar frequencies. For instance, among the relations we consider, four types form less than $0.5\%$ of the total edges in a task we consider. 
The edge relations are also very sparse: Less than $0.3\%$ of the ordered token pairs are connected (see Figure~\ref{fig:edge_on_freq_completion_large}). 
The sparsity of edges makes the binary classification problem for each edge type a label-imbalanced one, and we follow~\citet{johnson2020learning} in optimizing the focal loss~\cite{lin2017focal} as a remedy.

\subsection{The Edge Prediction Model}\label{subsec:edge_prediction_model}
\vspace{-2mm}
\paragraph{Architecture.}
Our edge prediction model is based on the Transformer base architecture without its final encoding-to-logit layer ($n_{\mathrm{layers}} =6$, $d_\mathrm{model}=512$, and $d_{\mathrm{ff}}=2048$). 
We add a final block to this architecture that produces a logit tensor which has size quadratic in the sequence length, each entry of which is used for computing a sigmoid score of whether a type of relation exists between an ordered token pair. 
This final block consists of $(i)$ a multi-head attention that produces a tensor of size $(B, L, L, H)$, followed by $(ii)$ a dense layer that produces a tensor of size $(B, L, L, E)$, where
$B$ is the batch size, $L$ is the sequence length, $H$ is the number of hidden units, and $E$ is the number of edge types. 
We use ``Pre-LN'' described in~\cite{xiong2020layer} and adopted in GPT-2~\cite{radford2019language}, where Layer Normalization~\cite{ba2016layer} is placed inside residual blocks. 
The multi-head attention in the final block has improved capacity with $H=32$ and $d_\text{model} = 1024$.
This architecture is similar to the dot product baseline used by~\citet{johnson2020learning}, but is different in that we train one model for all relation types. 

\vspace{-2mm}
\paragraph{Causal Masking.}
For applications with causal ordering (e.g. code completion), the model should not attend to a future token when producing the edge score for two earlier appearing tokens. This is because the model does not have access to future information at test time. 
Due to the edge prediction model being based on the Transformer, it suffices to apply the usual causal mask (masking out position $i$'s attention to any position $j > i$) to the base Transformer and the last attention block.
This is because the edge logit between any ordered pair of distinct positions $(i, j)$ is a position-wise non-linear transform of the dot product between the encodings $e_i$ and $e_j$, each of which is independent of future information after applying the causal mask.

\subsection{Using Predicted Relations}\label{sec:use_edge_prediction}
We apply the learned relation prediction model to code completion and localizing and repairing variable misuse. 
For the variable misuse task, we specifically focus on localizing and repairing such bugs for work-in-progress code that is broken.
This task is motivated by the desiderata that a machine-learning-powered IDE should be able to perform high quality analysis even on code that is temporarily broken due to intermittent editing by a developer.

The core architecture we adopt use relative position representations~\cite{shaw2018self,dai2019transformer,yang2019xlnet}. 
Specifically, in all Transformer architectures, we use the relative positional encoding described by~\citet{dai2019transformer}. 
The corresponding attention computation has the following form: 
\eq{
\label{eq:rel_attn}
\bm{\alpha}_{i, j} &=\underbrace{\mathbf{E}_{x_{i}}^{\top} \mathbf{W}_{q}^{\top} \mathbf{W}_{k, E} \mathbf{E}_{x_{j}}}_{(a)}+\underbrace{\mathbf{E}_{x_{i}}^{\top} \mathbf{W}_{q}^{\top} \mathbf{W}_{k, R} \mathbf{R}_{i-j}}_{(b)} +\underbrace{u^{\top} \mathbf{W}_{k, E} \mathbf{E}_{x_{j}}}_{(c)}+\underbrace{v^{\top} \mathbf{W}_{k, R} \mathbf{R}_{i-j}}_{(d)}.
}
Here, $\bm{\alpha}_{i,j}$ and $\mathbf{R}_{i-j}$ are respectively the attention score and relative positional encoding from the $i$th to the $j$th token in the sequence. 
$\mathbf{E}_{t}$ is the encoding for token $t$, $\mathbf{W}_q$ is used to compute queries, $\mathbf{W}_{k, E}$ is used to compute keys with the encoding, $\mathbf{W}_{k, R}$ is used to compute keys with the relative position, and $u$ and $v$ are biases shared across all multi-head attention layers. 
The choice of relative positional encoding as opposed to the absolute one is inspired by their improved performance on code summarization~\cite{2021languageagnostic}.

To leverage the additional edge relations provided by the relation prediction models, we associate each type of edge with a trainable embedding vector, and augment terms $(b)$ and $(d)$ in~\eqref{eq:rel_attn}. Concretely, suppose the ordered token pair $(x_i, x_j)$ has a relation of type $r$ with the embedding vector $\mathbf{R}^{(r)}$, the updated attention has the following form:
\eq{
\label{eq:rel_attn_pp}
    \!\widetilde{\bm{\alpha}}_{i,j} =
        \bm{\alpha_{i,j}} + 
        \underbrace{
            \mathbf{E}_{x_{i}}^{\top} \mathbf{W}_{q}^{\top} \mathbf{W}_{k, R'} \mathbf{R}^{(r)}
        }_{(b')} + 
        \underbrace{
            v^{\top} \mathbf{W}_{k, R'} \mathbf{R}^{(r)}
        }_{(d')}.
}
When there exist multiple relations between an ordered pair, we aggregate their contributions by summing the terms $(b')$ and $(d')$ for each edge type. 
For a multi-head attention layer, we replicate the update in~\eqref{eq:rel_attn_pp} across all heads.
With the aid of ground truth relations, $(b')$ has been used to modify the attention in the GREAT~\cite{hellendoorn2019global} and RAT-SQL~\cite{wang2019rat} models, whereas the combination of $(b')$ and $(d')$ has been used in the Code Transformer model~\cite{2021languageagnostic}.
Since the edges we model are sparse, the additional term in~\eqref{eq:rel_attn_pp} can be computed and backpropagated through with sparse primitives in standard automatic differentiation libraries. 
In practice, we observe a minor overhead for training\footnote{
We observe a $~7\%$ wall time training overhead (estimated using \texttt{torch.autograd.profiler}) of the relation-aware model compared to one without computing the additional term in~\eqref{eq:rel_attn_pp}. 
While wall time is dependent on the hardware and implementation, we expect to see similar numbers across similar settings. 
}. 

\subsubsection{Architecture for Code Completion}
The architecture for the code completion task is slightly involved due to the need to predict subwords, multiple of which form a token. 
This is at odds with the ground truth edge relations described in Section~\ref{subsec:ground_truth_relations}, which is between full tokens. 
While it is possible to extend the same edge relation between tokens for subwords in a fashion of creating bipartite graphs (connect every subword for token A to every subword for token B, if a relation exists between tokens A and B), 
our preliminary experiments suggest that learning such extended edges between subwords is difficult (e.g. precision and recall scores for certain edge types are lower than $0.6$ even after hyperparameter tuning). 
\begin{wrapfigure}[21]{R}{0.47\textwidth}
\vspace{-4mm}
\centering
{\includegraphics[width=0.45\textwidth]{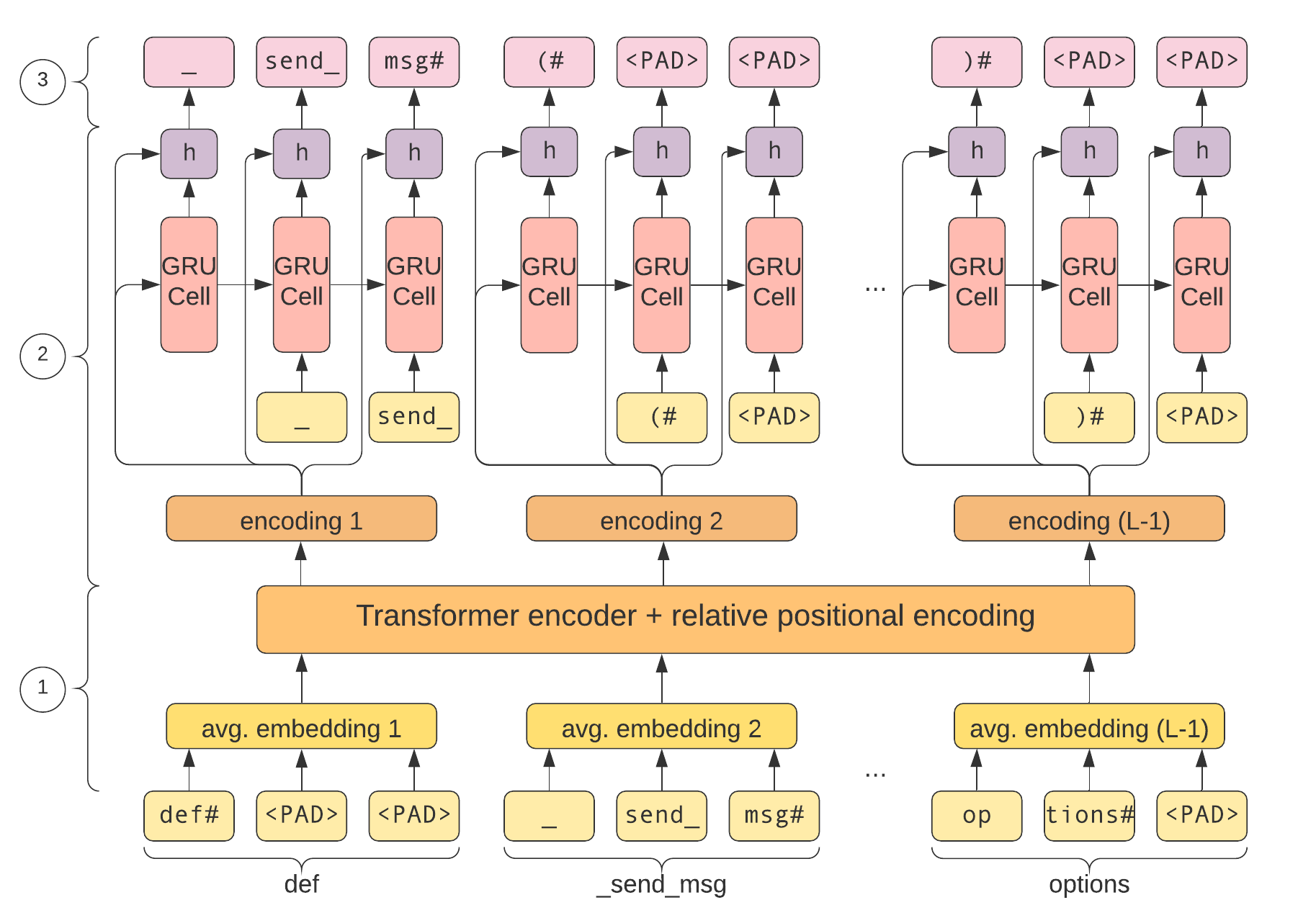}}
\caption{
Architecture for the code completion task consists of three blocks: 
\textcircled{1} a Transformer encoder with attention based on relative positions, \textcircled{2} an attention-based GRU, and \textcircled{3} a linear projection. 
The individual subtoken embeddings passed to \textcircled{1} is reused in \textcircled{2}.
For ease of demonstration, we limit each token to have $3$ subwords in the figure.
Hashtag marks the end of a token. 
}
\label{fig:completion_model}
\end{wrapfigure}

We propose to maintain the edge relation structure, but adjust the model for code completion to better adapt to the edge data. 
Specifically, to predict each token, our model predicts an encoding and generates all subwords for the token in an autoregressive manner conditional on the encoding.
Our architecture has three components: \textcircled{1} a Transformer encoder that takes in averaged embeddings of subwords for each token with attention scores computed based on~\eqref{eq:rel_attn_pp}, \textcircled{2} an attention-based Gated Recurrent Unit (GRU, \cite{cho2014learning}) decoder which predicts the next subword for each token given all previous subwords, and \textcircled{3} a linear projection layer that maps encodings to logits. See Figure~\ref{fig:completion_model} for an illustration.

The attention mechanism of the GRU \textcircled{2} is based on that described by~\citet{luong2015effective}: The subword embedding and hidden state from previous steps are given to the GRU cell to produce an output, which is concatenated with the encoding of the whole token produced by the Transformer. The joint encoding is passed through a Multi-Layer Perceptron (MLP) with a single hidden layer to obtain logits. 
We use a dimension size $1/2$ of that of the Transformer encoding for the hidden layer in the MLP. 
The overall model has roughly $2.3\%$ fewer parameters than the base Transformer due to this bottleneck hidden layer in the MLP. 
Since code completion has a natural temporal ordering, we apply causal masks at the token level during training to prevent the model from peaking at the future.

\subsubsection{Architecture for Variable Misuse}
For the variable misuse task, we adopt the model used by~\citet{hellendoorn2019global}, except that we compute attention scores using the formulation based on relative positions in~\eqref{eq:rel_attn_pp}.
In summary, the model takes in a sequences of subwords, averages the subword embeddings for each token, and feeds the embedding into a Transformer.
To produce candidate locations for bug and repair targets, the model includes a linear projection at the end which outputs two logits for each location. 

\subsection{Backpropagating Through Predicted Relations}\label{subsec:backprop_edge}
While a fixed relation model brings some performance gains (as we show in Section~\ref{sec:experiments}), the paradigm is not restricted to use only a fixed model. 
In fact, when training to optimize performance for a specific task, one may adapt the relation model by backpropagating through the discrete structure using any method within the established suite~\cite{williams1992simple,maddison2016concrete,jang2016categorical,bengio2013estimating,paulus2020gradient}. 

For simplicity, we explore using a variant of the straight-through estimator~\cite{hinton2012neural}. 
Specifically, for a binary relation random variable with logit $l$, we use the ``hard-version'' $\mathbbm{1}[l \ge 0]$ during the forward pass, and during backpropagation treat the variable as if the ``soft-version'' $\sigma(l / \tau)$ were used in the forward pass. 
Here, $\sigma(\cdot)$ is the sigmoid function and $\tau$ is a temperature hyperparameter.
This node operates differently than the Gumbel-Softmax/Concrete random variable, since no additional noise variable is sampled. 
While~\citet{bengio2013estimating} reported worse results when using this estimator compared to straight-through, we found the latter to fail on our task. 

We will see that even on clean code, fine-tuning the relation prediction model in this manner leads to improved performance compared to just using the fixed ground truth edges. 
This improvement is also observed across different settings for work-in-progress code with learned edges (see Section~\ref{subsec:finetune_edge}).

\section{Experiments}\label{sec:experiments}
We present experiments in four subsections. 
We first verify that we can learn edge prediction models of reasonable quality across a suite of edge types. 
We then present results on two applications: $(a)$ code completion without assuming the existence of any partial AST structure, and $(b)$ localizing and repairing variable misuse for ill-formed work-in-progress code.
To close the section, we present some analyses for fine-tuning the edge prediction model during task optimization. 
All reported numbers are averaged over three seeds. Error bars are based on one standard deviation.

\subsection{Learning Relations for Source Code}\label{subsec:learn_edges}
We show that the edge prediction model described in Section~\ref{subsec:edge_prediction_model} is able to learn a wide variety of relations by training such as models for each task separately on their task specific data for the two tasks we consider (code completion and variable misuse).
We report the shared settings and overall results in this subsection. 

We trained all edge prediction models with the Adam optimizer~\cite{kingma2014adam} using a fixed learning rate of $0.0001$ and a batch size of $48$. For other hyperparameters, we adopted the default in \texttt{PyTorch}. 
We performed early stopping based on validation performance with a patience of $50$k iterations. 
For focal loss, we set its hyperparameter $\gamma=2$ for all experiments as this was reported to be near optimal for a wide array of settings~\cite{lin2017focal}.
Training the edge prediction model is memory intensive due to its last layer instantiating several tensors with size quadratic in the sequence length $L$ and linear in the number of hidden units $H$ or edge types $E$. 
To reduce the memory footprint during training, we apply gradient checkpointing for the multi-head attention layers. 
With this setup, training converges to a validation F score of around $99\%$ within $3$ days for both tasks using a single V100 GPU. 

\begin{figure*}[t]
\begin{minipage}[t]{0.325\linewidth}
\centering
{\includegraphics[width=0.98\textwidth]{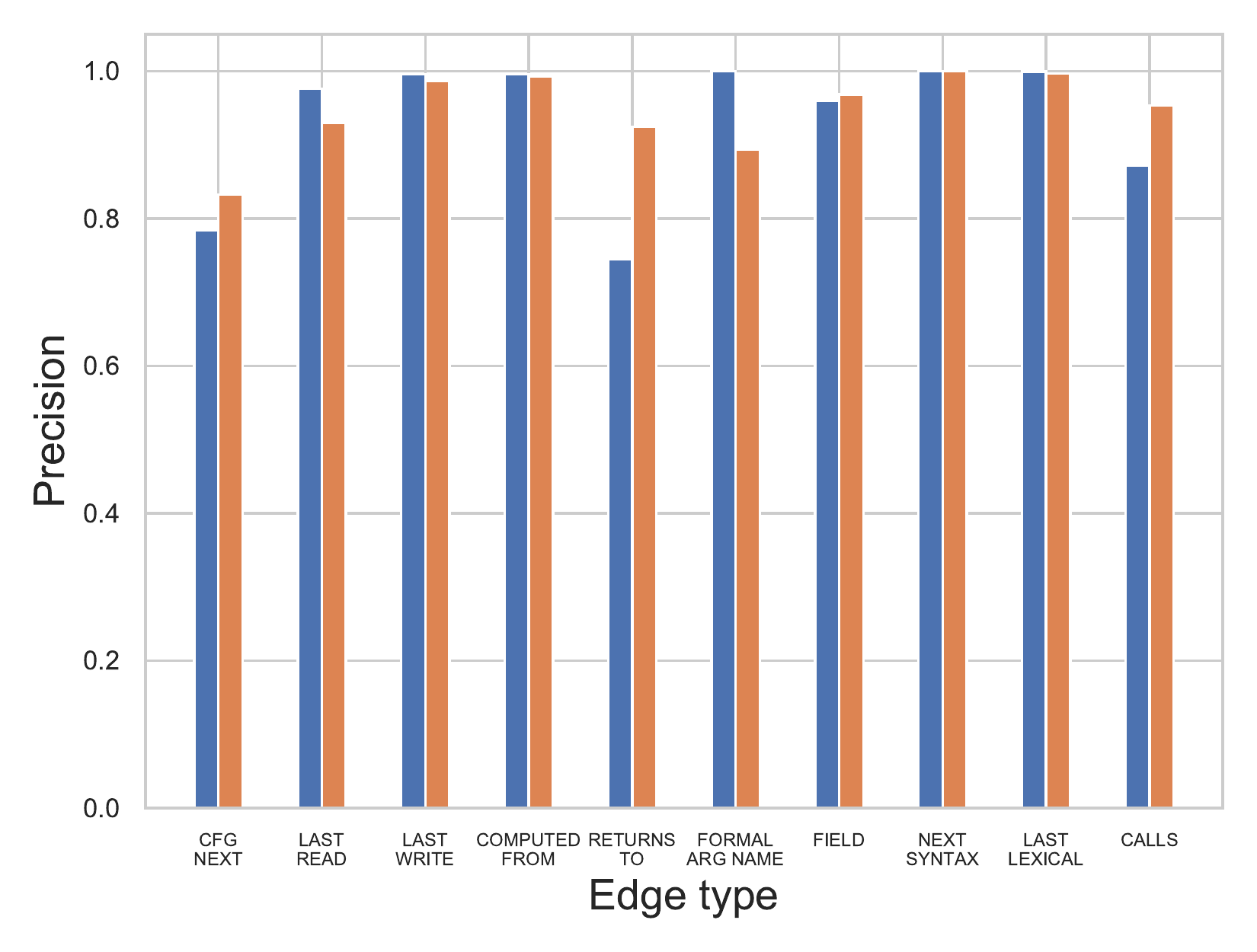}}
\\  (a) precision
\end{minipage}
\begin{minipage}[t]{0.325\linewidth}
\centering
{\includegraphics[width=0.98\textwidth]{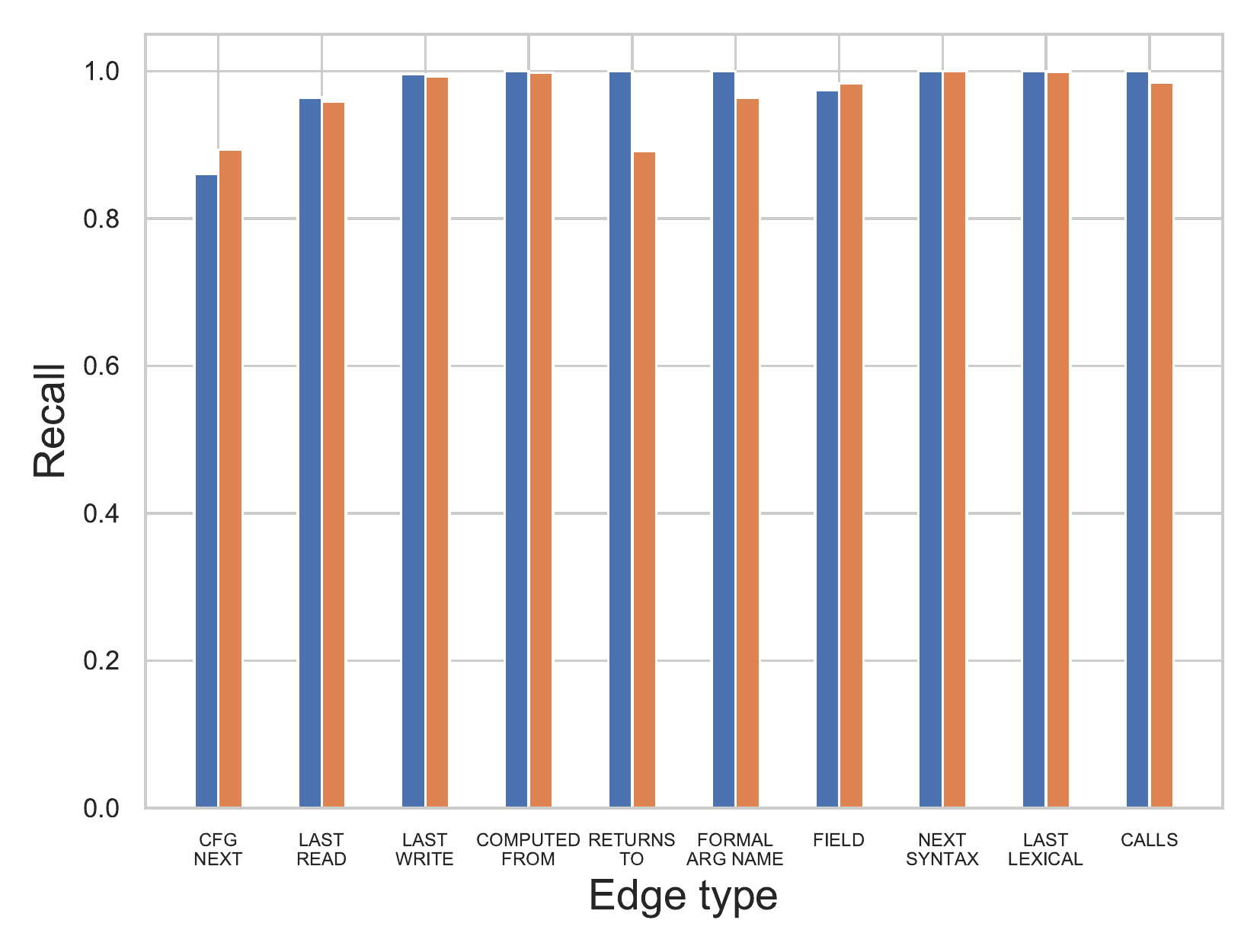}}
\\  (b) recall
\end{minipage}
\begin{minipage}[t]{0.325\linewidth}
\centering
{\includegraphics[width=0.98\textwidth]{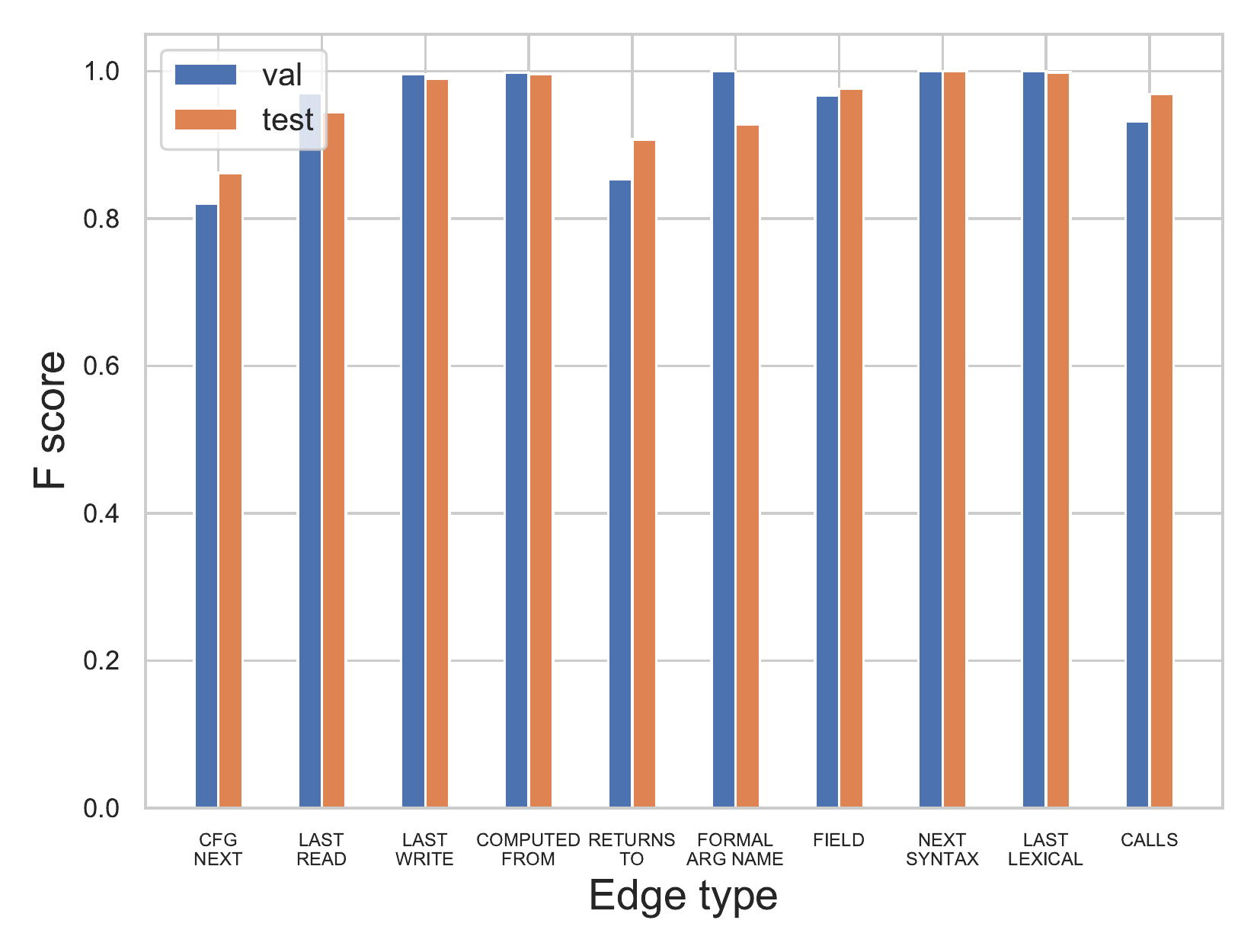}}
\\  (c) F score
\end{minipage}
\caption{
Relation prediction model produces high quality edges on the validation and test splits of the Python corpus~\cite{karampatsis2020big}. 
The model has an F score above $0.8$ for all edge types, and attains an F score close to $1.0$ for the most frequently occurring types.
}
\vspace{-2mm}
\label{fig:edge_prediction_completion}
\end{figure*}

\paragraph{Results.}
Since occurrences of edges are extremely sparse, a model that always predicts the non-existence of edges can attain an accuracy beyond $99\%$ on the test split. 
We therefore report the precision, recall, and F score. 
Despite having a simple architecture, the edge prediction model is able to produce high quality predictions for the code completion task and attains an F score close to $1.0$ for edge types with the highest frequencies (see Figure~\ref{fig:edge_prediction_completion}). 
For low frequency edge types and edge types related to control flow (e.g. ReturnsTo and CFGNext), the model performs slightly worse, but still achieves an F score beyond $0.8$. 
Similar but slightly worse results are obtained for the variable misuse task; see \ref{app:edge_prediction_varmisuse} for results.

\subsection{Code Completion}\label{subsec:code_completion}
We adopted the Python source code dataset used by~\citet{karampatsis2019maybe} that is comprised of two training splits (one small and one large), a validation, a test, and an encoding split used only for learning a tokenization. The corpus sanitized by~\citet{karampatsis2020big} is licensed under CC BY 4.0. 

\begin{table}[t]
\footnotesize
\setlength\tabcolsep{2.4pt}
\caption{
Results for next token prediction. 
Small and Large respectively refer to when the small and large training split is used. 
Numbers in bold are the best results in each column.
We exclude comparing to Aware-true which cannot be obtained in practice given any unseen code prefix. 
}
\centering
\begin{tabular}{c cccc}
\toprule
\multirow{2}[2]{*}{Model} & \multicolumn{2}{c}{\text{Perplexity ($\downarrow$)}}
& \multicolumn{2}{c}{\text{Top-1 accuracy in $\%$ ($\uparrow$)}} \\
\cmidrule(lr){2-3}
\cmidrule(lr){4-5}
 & Small & Large & Small & Large \\
\midrule
{\footnotesize Transformer (stride=100)}  & $12.084 \pm 0.215$& $4.249 \pm 0.051$& $66.152 \pm 0.583$& $76.380 \pm 0.188$\\ 
{\footnotesize Transformer (stride=5)}   & $\bm{12.002 \pm 0.221}$& $4.217 \pm 0.051$& $66.158 \pm 0.584$& $76.444 \pm 0.189$\\ 
\midrule 
Agnostic   & $14.858 \pm 0.170$& $3.764 \pm 0.019$& $71.576 \pm 0.106$& $79.776 \pm 0.064$\\ 
Aware-fixed   & $14.637 \pm 0.045$& $3.782 \pm 0.012$& $71.857 \pm 0.156$& $79.686 \pm 0.002$\\ 
Aware-tuned   & $12.381 \pm 0.189$& $\bm{3.495 \pm 0.016}$& $\bm{72.350 \pm 0.065}$& $\bm{80.745 \pm 0.059}$\\ 
\midrule
Aware-true   & $13.458 \pm 0.171$& $3.535 \pm 0.003$& $73.627 \pm 0.148$& $81.347 \pm 0.017$ \\
\bottomrule
\end{tabular}
\label{table:completion_token_level}
\end{table}

\paragraph{Dataset Preprocessing.}
We followed the preprocessing setups in~\cite{hellendoorn2017deep,karampatsis2019maybe}: replace non-ASCII character sequences such as Chinese ideograms inside strings with the special token \textlangle non-en\textrangle, remove comments, and replace string literals consisting of 15 characters or more by the empty string. 
After preprocessing, the small training, large training, validation and test sets contain $20738$, $999000$, $14090$ and $12865$ files, respectively.
Since the dataset consists of Python projects mined directly from Github, certain sequences are very long and contain a variety of long words.
We also converted all files in Python2 to Python3 using Python's automated translation tool \texttt{lib2to3}. 
To reduce the size of the vocabulary, we followed~\citet{karampatsis2020big} in splitting tokens into subtokens with the byte-pair encoding algorithm~\cite{sennrich2015neural} ran for 10k merges using the \texttt{Tokenizer} package from \texttt{Hugging Face}.
While the model presented in Section~\ref{sec:use_edge_prediction} is in principle able to handle tokens that consist of an arbitrary number of subwords, we truncate each token to have a maximum of $6$ subwords for computational efficiency. 
This covers $98.9\%$ and $99.2\%$ of the vocabulary in the small training and validation sets, respectively.
We trained the edge model with the usual causal masks for Transformers. 

\paragraph{Training and Evaluation.}
Since many files are longer than the typical context size of a Transformer-based model, which parts of a file should one train with becomes a question. 
Unlike~\citet{karampatsis2020big}, we did not take a fixed size prefix of each file.
Instead, we sampled uniformly a new window each time a file is selected for a gradient update. 
The window is of the context size if the file is longer; otherwise, it is chosen to be the whole file. 
For testing, we sampled $3$ fixed windows for each test file. 
We set the context size to be $256$ for computational efficiency.

We compared the performance of the relation-aware architecture described in Section~\ref{sec:use_edge_prediction} trained and tested with a learned edge prediction model against a model with no relational information (relation-agnostic).
We also report results for the relation-aware models trained and tested with ground truth edges as a guideline for the gain one could expect with the present suite of edge types. 

As a baseline, we also report results for the Transformer base architecture with attention scores computed as in~\eqref{eq:rel_attn}, modeling sequences directly at the subword level\footnote{The architecture is therefore very similar to Transformer-XL~\cite{dai2019transformer} with the minor difference being the positioning of Layer Normalization and Dropout. 
Since the present focus is not long sequence modeling, our models also don't have their segment-level recurrence.
}. 
During training, we sampled windows of subwords of context size $256$. 
Since examples in the test set can vary in length and contain up to $256$ tokens, which may be ``flattened'' into much more than $256$ subwords, we performed the typical sliding window evaluation starting at the beginning of each test window and report results with various strides. 
With our implementation, each gradient update for the baseline is roughly $1.7$x as fast as that for the relation-aware architecture proposed in Section~\ref{sec:use_edge_prediction} in terms of wall time. 
However, inference for the baseline is much slower due to the sliding window evaluation (approximately $1/10$ and $1/20$ the speed of relation-aware model with strides $100$ and $5$, respectively). 

For all experiments, we used the Adam optimizer with a fixed learning rate of $0.0001$ and adopted the $\texttt{PyTorch}$ default for other hyperparameters.
We applied gradient clipping with a max $2$-norm of $0.25$. 
These hyperparameter choices were primarily optimized for training stability. 
Additionally, we set the dropout rate to be $0.1$ and also randomly dropout subwords to reduce overfitting. We used a batch size of $32$ and data parallel training across $8$ V100 GPUs across all settings. 
We trained all models for at most $1$k epochs and early stopped based on validation performance with a patience of $50$k updates. 
Training time is between $1$-$2$ weeks for the large dataset and less than $1$ day for the small dataset.

\paragraph{Results.}
We report the perplexity and top-1 accuracy for next token prediction in Table~\ref{table:completion_token_level}. 
The table shows that compared to a relation-agnostic model, the relation-aware model given the ground truth edges is able to perform much better in both perplexity and accuracy. 
Nevertheless, when using only learned and fixed edges, the relation-aware model performs similarly as the relation-agnostic baseline. 
Further fine-tuning the edge prediction model brings a noticeable performance gain to the relation-aware setting ($0.774\%$ and $0.969\%$ absolute gain in top-1 accuracy for Small and Large). 

Notably, both the agnostic and aware model perform better than the baseline transformers which directly model subword sequences on all metrics except the perplexity score when the small training split is used.
Results on next subword prediction have the same trend. See~\ref{app:completion_subword} for details.

To diagnose the poor performance of the relation-aware model supplied with learned and fixed edges, we conducted ablation studies and observe that the edges types most difficult to predict are the most influential in achieving the performance gain. 
This indicates that downstream accuracy is non-uniformly affected by
different edge types and suggests that future work may shift
edge prediction capacity to the harder edge types. 
Details of this study can be found in \ref{app:completion_ablation}.

We do not compare directly against results in~\cite{karampatsis2020big}, since our setting is not strictly comparable to theirs due to subtle differences in data preprocessing (e.g. the specific procedure of creating subwords is different).
However, we note that our reported perplexities are in the range of that reported for their static evaluation setting and is lower possibly due to using the Transformer architecture and that we don't merely train on file prefixes (e.g. our models trained on the large split attains a per-token cross entropy around $1.2$, whereas they report a cross entropy of $2.09$ for their LSTM models). 

\subsection{\small Localizing and Fixing Variable Misuse  for WIP Code}
\begin{table}[t]
\caption{
\footnotesize
Results for the variable misuse task on ill-formed code each with $k$ corruptions. 
}
\footnotesize
\setlength\tabcolsep{2.4pt}
\centering
\begin{tabular}{c c ccc ccc}
\toprule 
\multirow{2}[2]{*}{Task} & \multirow{2}[2]{*}{Model} & \multicolumn{3}{c}{\text{Small}} & \multicolumn{3}{c}{\text{Large}} \\
\cmidrule(lr){3-5} \cmidrule(lr){6-8} & & $k=1$ & $k=2$ & $k=5$ & $k=1$ & $k=2$ & $k=5$ \\
\midrule
\multirow{3}{*}{Localization} & Agnostic & $44.0\pm0.8$ & $43.0\pm1.7$ & $42.3\pm2.2$ & $75.4\pm1.1$ & $76.0\pm0.3$ & $72.6\pm1.1$ \\
& Aware-fixed & $61.5\pm0.5$ & $61.6\pm0.1$ & $60.6\pm0.5$ & $76.6\pm0.3$ & $75.6\pm0.6$ & $73.3\pm1.7$ \\
& Aware-tuned & $\bf{68.6\pm0.2}$ & $\bf{68.0\pm0.4}$ & $\bf{67.6\pm0.4}$ & $\bf{79.2\pm0.1}$ & $\bf{79.0\pm0.1}$ & $\bf{78.4\pm0.3}$ \\
\midrule
\multirow{3}{*}{Repair} & Agnostic & $25.9\pm1.1$ & $24.0\pm2.1$ & $23.4\pm2.7$ & $63.7\pm6.3$ & $69.6\pm1.1$ & $65.5\pm0.8$ \\
& Aware-fixed & $45.0\pm1.0$ & $44.1\pm1.2$ & $42.5\pm0.3$ & $69.3\pm0.3$ & $66.4\pm4.6$ & $66.7\pm1.6$ \\
& Aware-tuned & $\bf{55.8\pm0.2}$ & $\bf{55.7\pm0.3}$ & $\bf{54.4\pm0.1}$ & $\bf{73.9\pm0.5}$ & $\bf{73.8\pm0.0}$ & $\bf{72.8\pm0.2}$ \\
\bottomrule
\end{tabular}
\vspace{-4mm}
\label{table:corrupted_varmisuse_repair}
\end{table}

With a pretrained edge prediction model, we show that relational structure can be generalized to ill-formed code to improve task performance. 
We focus on a setting where the code to be analyzed for variable misuse issues has undergone multiple mild perturbations such that a standard parser would fail to generate ASTs\footnote{While it is desirable to seek for methods that may handle strong perturbations, learning to generalize to out-of-distribution examples ``in the wild'' is still in general an open problem in machine learning~\cite{koh2020wilds}. 
}.
We consider the following set of perturbations:
\begin{itemize}[topsep=0pt,leftmargin=*]
\setlength\itemsep{0.05em}
    \item Keyword: Randomly corrupt keywords in the language (e.g. \texttt{for}, \texttt{while}, \texttt{if}, \texttt{def} in \texttt{Python}). 
    \item Deletion: Randomly delete a fixed number of tokens. 
    \item Punctuation: Add random punctuation marks at random locations. 
    \item Indentation\footnote{
    We avoid (re)indenting only the first line in any indentation block, since in this scenario, any standard tokenizer would be confused about the amount of whitespace characters for each tab/indentation.
    }: Indent or dedent a randomly selected span of lines. 
\end{itemize}
We based our dataset on that used by~\citet{hellendoorn2019global} and applied a fixed amount of perturbations to each file. 
The perturbations were uniformly sampled to be one of the types described above.
We created three perturbed datasets, varying the number of corruptions applied to each example in the training, validation, and test splits (number of corruptions $k=1,2,5$).
To show the effect of dataset size on the model's performance, we created a separate small training set based on subsampling $1\%$ of the original training set. 
For efficiency, we also ignored examples in the training and test sets that are longer than the context size of $512$ tokens. 
We adopted the same hyperparameters and setup for experiments in this section as those used previously, except we parallelize training across $4$ V100 GPUs for each experiment. 
Training time is between $1$-$2$ weeks for the large training set and less than $1$ day for the small training set. 

\vspace{-2mm}
\paragraph{Results.}
Table~\ref{table:corrupted_varmisuse_repair} shows that with a fixed pretrained edge prediction model, a relation-aware model (Aware-fixed) is able to achieve a much better localization and repair performance compared to the relation-agnostic baseline (Agnostic) when the training set is small. 
This advantage, however, diminishes as training data become abundant. 
On the other hand, by further fine-tuning the edge prediction model during downstream learning (Aware-tuned), the relation-aware model achieves consistent gains compared to both using the learned-but-fixed edges and the relation-agnostic baseline.

\subsection{Fine-tuning the Relation Prediction Model: Is Pretraining to Predict Edges Necessary? }\label{subsec:finetune_edge}
We aim to gain a better understanding of why fine-tuning the relation prediction model helps. 
Specifically, we seek to answer the following question: Is fine-tuning the relation model helpful solely because the joint model has more parameters or is it that the joint model is benefitting from the structure learned by the relation model? 
To answer this question, we trained a relation-aware model for the variable misuse task where edges come from a randomly initialized relation model that is jointly optimized with the task model ({aware-random-init}). 
We compare this model against 
the relation-aware model with fine-tuned edges initialized from pretraining ({aware-pretrain-init}), and the relation-agnostic model ({agnostic}) for both well-formed and work-in-progress code.

\begin{figure}[t]
\begin{minipage}[t]{0.49\linewidth}
\centering
{\includegraphics[width=0.8\textwidth]{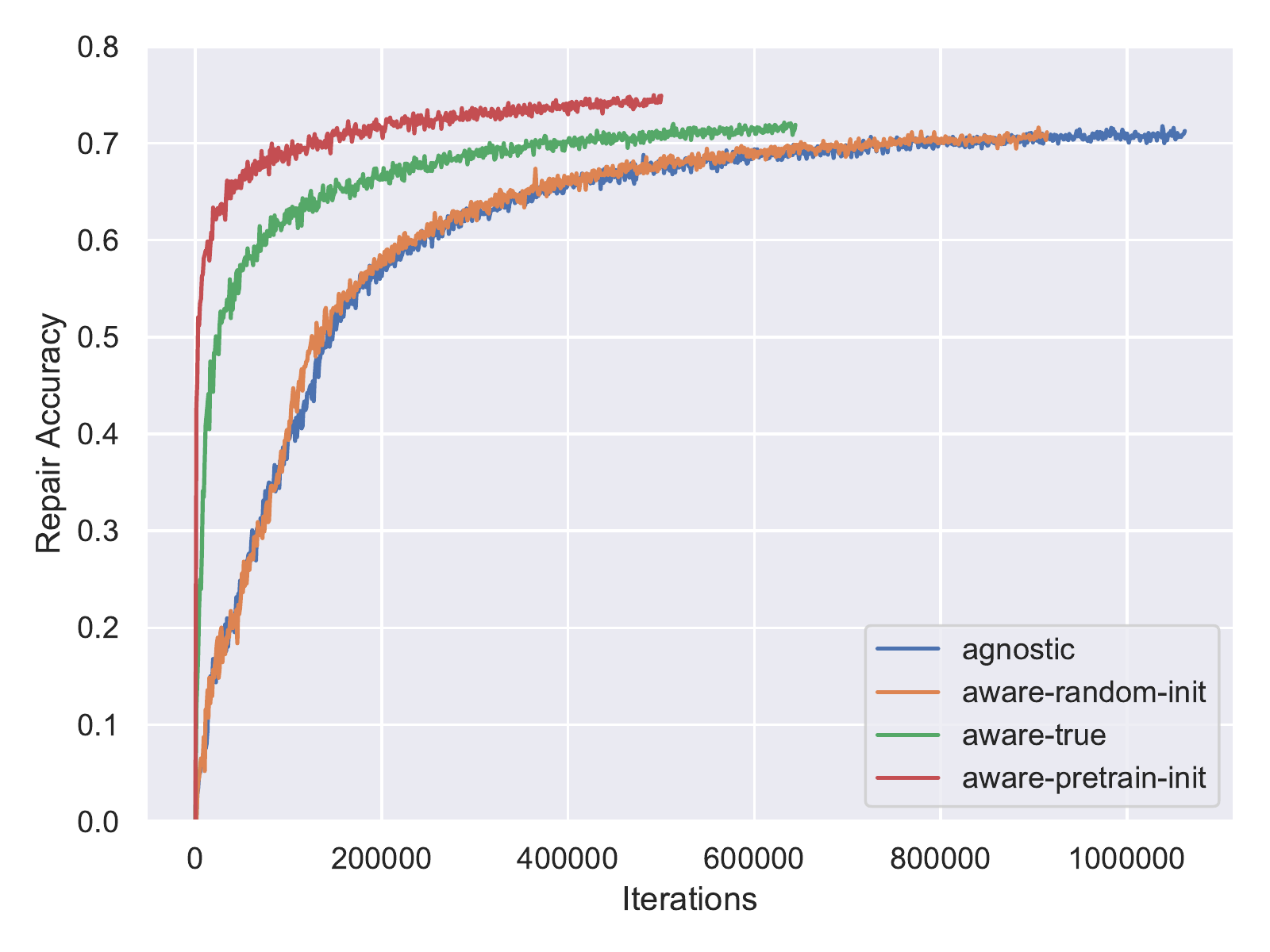}}
\\ \vspace{-1mm} (a) well-formed
\end{minipage}
\begin{minipage}[t]{0.49\linewidth}
\centering
{\includegraphics[width=0.8\textwidth]{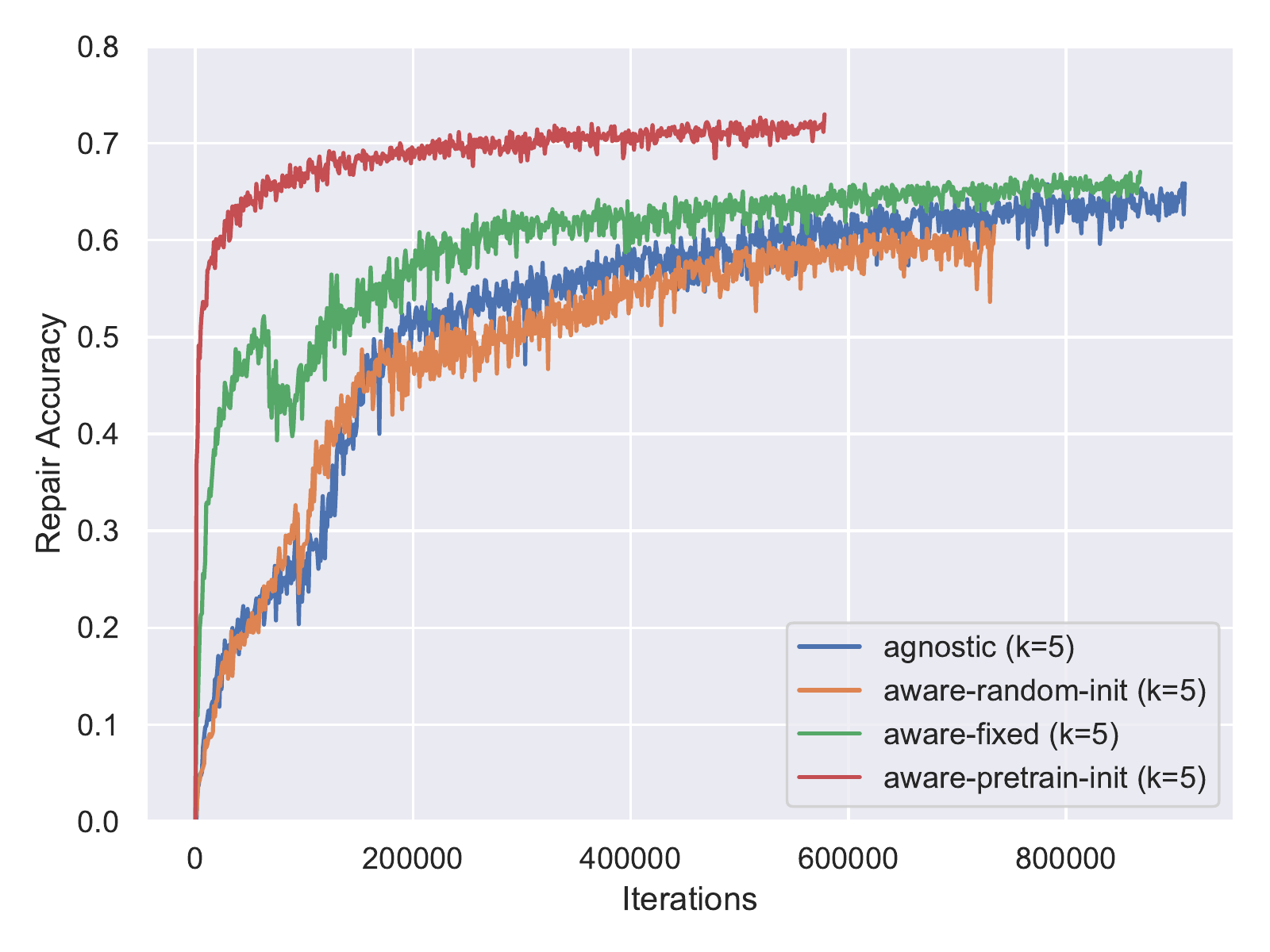}}
\\ \vspace{-1mm} (b) work-in-progress
\end{minipage}
\caption{
Fine-tuning the edge prediction model leads to improved repair accuracy on well-formed and work-progress code. 
Different curves terminate at different times due to early stopping.
Work-in-progress examples in (b) each contain $k=5$ corruptions. 
``aware-fixed'' uses a pretrained relation model that is not jointly optimized with the task model. 
``aware-true'' uses the ground truth edges and is only applicable for well-formed code. 
Results are based on training on the large training splits. 
}
\label{fig:clean_and_wip_finetune}
\end{figure}

\vspace{-2mm}
\paragraph{Results.}
Figure~\ref{fig:clean_and_wip_finetune} (a) demonstrates that fine-tuning a pretrained relation model during downstream learning improves on using the ground truth edges\footnote{
The careful reader who champions the utility of hand-crafted relations derived from expert knowledge might find this result curious. 
We note it is in general unsurprising that features extracted from large amounts of data with neural nets may beat their hand-crafted counterparts in the data-rich regime~\cite{devlin2018bert,krizhevsky2012imagenet,radford2018improving}
}. 
Moreover, jointly training relation-aware models with relation models initialized with random weights has a performance no better than the relation-agnostic baseline. 
This suggests that the pretrain-then-fine-tune combination is the key to producing a relation model useful for downstream learning. 
The same trend consistently holds for models trained on the small training split and work-in-progress code with various perturbation levels (see~\ref{app:finetune} for results and comments).

\section{Related Work}
The problem of identifying structure from broken and incomplete code is not new. 
In the programming languages community, the problem has been studied as part of robust parsing and error handling in compilers~\cite{weir1992efficiency,aho1986compilers}, where popular approaches are based on parsing with non-standard productions~\cite{moonen2001generating,nilsson2008practical,van1999building,moonen2002lightweight}. 
These approaches rely on language dependent and hand-designed grammars and typically require multiple stages of execution.
Notably, the fuzzy parser in Joern runs in three stages, each of which follows a separate bespoke island grammar~\cite{yamaguchi2015pattern}.
While being less structured than robust parsing, our approach of learning and transferring edges is more flexible and direct.

We draw inspiration from a number of recent works on learning edge structure for use
in relation-aware neural networks and graph neural networks \cite{johnson2016learning,kipf2018neural,franceschi2019learning,johnson2020learning}.
Compared to these works, the main methodological distinction comes from our assumption about the availability of strong supervision.
Whereas strong supervision for latent variables is often costly to obtain, e.g., meaning representations in semantic parsing \cite{yin2018structvae}, 
our key observation is that it is readily available for edge structures in well-formed code.
Thus, the main challenge is transferring from the distribution where strong supervision is available to a distribution where it is not.
Perhaps the most similar work is that by~\citet{velickovic2020pointer}, which assumes strong supervision for edges is available on small examples, but then the goal is to systematically generalize to longer examples. 
A difference is that in our setting, there is not clearly the same kind of systematicity in the needed generalization.

From the application perspective, there are alternative approaches to dealing with work-in-progress code.
A common approach for source code modeling is to define relations that can be computed from a program prefix or a partial expansion using standard productions \cite{maddison2014structured,nguyen2015graph,parisotto2016neuro,rabinovich2017abstract,yin2017syntactic,brockschmidt2018generative,kim2020code,alon2020structural}.
Relations are then explicitly defined for the work-in-progress case and can be computed deterministically. However, these approaches usually are only possible in restricted cases and would not be applicable to all useful edge types, or remain valid for all corruptions encountered in practice.

It is worthwhile to mention that to handle slightly buggy code, a different approach would be to leverage existing bug fixers~\cite{dinella2020hoppity,bader2019getafix,sadowski2015tricorder} to transform buggy instances into well-formed ones and thereafter perform downstream tasks based on the latter as input. 
While potentially excelling at handling code with common errors~\cite{bader2019getafix}, this approach is limited in handling the incomplete code prefixes studied in this paper. 
Aside, we note that the approach is not inherently in conflict with ours, as one could imagine leveraging learned relations to build improved neural bug fixers.  

While our work is not solely devoted to code completion, we comment on several recent neural-based attempts in the literature.
\citet{svyatkovskoy2020fast} consider aiding completion candidate providers (most notably ones based on static analysis) with learned neural representations. 
Their setting differs from ours in assuming access to a candidate provider of reasonable quality upfront. 
On the other hand, \citet{svyatkovskiy2020intellicode} and \citet{wang2020towards} respectively study multilingual and whole line completion using Transformers.
Our work is, in every respect, orthogonal to the two aforementioned, as the idea of leveraging fine-tuned relations for better completion is applicable to both settings

Lastly, we comment that ML4Code researchers have borrowed successful ideas in the NLP community such as pretraining large Transformers on heterogenous datasets for transfer learning and multi-task learning~\cite{DBLP:journals/corr/abs-2002-08155,DBLP:journals/corr/abs-2009-08366,DBLP:journals/corr/abs-2104-02443,kanade2020learning}.
The idea of learning to predict edges has already seen modest successes for pretraining~\citep{DBLP:journals/corr/abs-2009-08366}.
Exploring whether learning relations could be useful beyond this setting is an interesting direction for future research.

\section{Conclusion and Limitation}
We showed that by simply learning a relation prediction model, the notion of program graphs can be generalized to work-in-progress code. 
This enables relation-aware models to be used for source code that may be broken or incomplete. 
Task performance is significantly improved when a pretrained relation model is simultaneously fine-tuned during task optimization for both work-in-progress and well-formed code. 
Future work may investigate whether the fine-tuning approach could improve the performance on other tasks and with other types of relational structure (such as ones used by graph neural networks for algorithmic reasoning~\cite{velickovic2020pointer}, ones in text data with corresponding parse trees~\cite{nivre2016universal,nivre2020universal}, or even ones in knowledge graphs~\cite{ehrlinger2016towards}). 

Having improved flexibility and capacity, modern-day neural-based approaches for learning are typically at the same time less transparent and interpretable.
Our relation learning procedure is no outlier in this regard: Fine-tuned relation models may predict edges that are unexpected and conform less to their initially prescribed category. 
Future work may improve the explainability and interpretability aspects of adapted relations. 

\section*{Acknowledgements}
We thank Michael Sloan and Tatsunori Hashimoto for helpful discussions.
We thank the ML4Code team at Google and Michihiro Yasunaga for helpful feedback on an earlier draft.

\bibliography{main}

\begin{thebibliography}{84}
\providecommand{\natexlab}[1]{#1}
\providecommand{\url}[1]{\texttt{#1}}
\expandafter\ifx\csname urlstyle\endcsname\relax
  \providecommand{\doi}[1]{doi: #1}\else
  \providecommand{\doi}{doi: \begingroup \urlstyle{rm}\Url}\fi

\bibitem[joe()]{joern}
Joern: Open-source code analysis platform for c/c++/java based on code property
  graphs.
\newblock \url{https://github.com/joernio/joern}.
\newblock Accessed: 2021-05-20.

\bibitem[Aho et~al.(1986)Aho, Sethi, and Ullman]{aho1986compilers}
Alfred~V Aho, Ravi Sethi, and Jeffrey~D Ullman.
\newblock Compilers, principles, techniques.
\newblock \emph{Addison wesley}, 7\penalty0 (8):\penalty0 9, 1986.

\bibitem[Allamanis et~al.(2017)Allamanis, Brockschmidt, and
  Khademi]{allamanis2017learning}
Miltiadis Allamanis, Marc Brockschmidt, and Mahmoud Khademi.
\newblock Learning to represent programs with graphs.
\newblock \emph{arXiv preprint arXiv:1711.00740}, 2017.

\bibitem[Allamanis et~al.(2018)Allamanis, Barr, Devanbu, and
  Sutton]{allamanis2018survey}
Miltiadis Allamanis, Earl~T Barr, Premkumar Devanbu, and Charles Sutton.
\newblock A survey of machine learning for big code and naturalness.
\newblock \emph{ACM Computing Surveys (CSUR)}, 51\penalty0 (4):\penalty0 1--37,
  2018.

\bibitem[Alon et~al.(2018)Alon, Brody, Levy, and Yahav]{alon2018code2seq}
Uri Alon, Shaked Brody, Omer Levy, and Eran Yahav.
\newblock code2seq: Generating sequences from structured representations of
  code.
\newblock \emph{arXiv preprint arXiv:1808.01400}, 2018.

\bibitem[Alon et~al.(2019)Alon, Zilberstein, Levy, and Yahav]{alon2019code2vec}
Uri Alon, Meital Zilberstein, Omer Levy, and Eran Yahav.
\newblock code2vec: Learning distributed representations of code.
\newblock \emph{Proceedings of the ACM on Programming Languages}, 3\penalty0
  (POPL):\penalty0 1--29, 2019.

\bibitem[Alon et~al.(2020)Alon, Sadaka, Levy, and Yahav]{alon2020structural}
Uri Alon, Roy Sadaka, Omer Levy, and Eran Yahav.
\newblock Structural language models of code.
\newblock In \emph{International Conference on Machine Learning}, pages
  245--256. PMLR, 2020.

\bibitem[Ba et~al.(2016)Ba, Kiros, and Hinton]{ba2016layer}
Jimmy~Lei Ba, Jamie~Ryan Kiros, and Geoffrey~E Hinton.
\newblock Layer normalization.
\newblock \emph{arXiv preprint arXiv:1607.06450}, 2016.

\bibitem[Bader et~al.(2019)Bader, Scott, Pradel, and Chandra]{bader2019getafix}
Johannes Bader, Andrew Scott, Michael Pradel, and Satish Chandra.
\newblock Getafix: Learning to fix bugs automatically.
\newblock \emph{Proceedings of the ACM on Programming Languages}, 3\penalty0
  (OOPSLA):\penalty0 1--27, 2019.

\bibitem[Ben-Nun et~al.(2018)Ben-Nun, Jakobovits, and Hoefler]{ben2018neural}
Tal Ben-Nun, Alice~Shoshana Jakobovits, and Torsten Hoefler.
\newblock Neural code comprehension: A learnable representation of code
  semantics.
\newblock \emph{arXiv preprint arXiv:1806.07336}, 2018.

\bibitem[Bengio et~al.(2013)Bengio, L{\'e}onard, and
  Courville]{bengio2013estimating}
Yoshua Bengio, Nicholas L{\'e}onard, and Aaron Courville.
\newblock Estimating or propagating gradients through stochastic neurons for
  conditional computation.
\newblock \emph{arXiv preprint arXiv:1308.3432}, 2013.

\bibitem[Bieber et~al.(2020)Bieber, Sutton, Larochelle, and
  Tarlow]{bieber2020learning}
David Bieber, Charles Sutton, Hugo Larochelle, and Daniel Tarlow.
\newblock Learning to execute programs with instruction pointer attention graph
  neural networks.
\newblock \emph{arXiv preprint arXiv:2010.12621}, 2020.

\bibitem[Brockschmidt et~al.(2018)Brockschmidt, Allamanis, Gaunt, and
  Polozov]{brockschmidt2018generative}
Marc Brockschmidt, Miltiadis Allamanis, Alexander~L Gaunt, and Oleksandr
  Polozov.
\newblock Generative code modeling with graphs.
\newblock \emph{arXiv preprint arXiv:1805.08490}, 2018.

\bibitem[Chen et~al.(2018)Chen, Liu, and Song]{chen2018tree}
Xinyun Chen, Chang Liu, and Dawn Song.
\newblock Tree-to-tree neural networks for program translation.
\newblock \emph{Advances in neural information processing systems},
  31:\penalty0 2547--2557, 2018.

\bibitem[Cho et~al.(2014)Cho, Van~Merri{\"e}nboer, Gulcehre, Bahdanau,
  Bougares, Schwenk, and Bengio]{cho2014learning}
Kyunghyun Cho, Bart Van~Merri{\"e}nboer, Caglar Gulcehre, Dzmitry Bahdanau,
  Fethi Bougares, Holger Schwenk, and Yoshua Bengio.
\newblock Learning phrase representations using rnn encoder-decoder for
  statistical machine translation.
\newblock \emph{arXiv preprint arXiv:1406.1078}, 2014.

\bibitem[Cousot and Cousot(1977)]{cousot1977abstract}
Patrick Cousot and Radhia Cousot.
\newblock Abstract interpretation: a unified lattice model for static analysis
  of programs by construction or approximation of fixpoints.
\newblock In \emph{Proceedings of the 4th ACM SIGACT-SIGPLAN symposium on
  Principles of programming languages}, pages 238--252, 1977.

\bibitem[Cummins et~al.(2020)Cummins, Fisches, Ben-Nun, Hoefler, and
  Leather]{cummins2020programl}
Chris Cummins, Zacharias~V Fisches, Tal Ben-Nun, Torsten Hoefler, and Hugh
  Leather.
\newblock Programl: Graph-based deep learning for program optimization and
  analysis.
\newblock \emph{arXiv preprint arXiv:2003.10536}, 2020.

\bibitem[Cvitkovic et~al.(2019)Cvitkovic, Singh, and
  Anandkumar]{cvitkovic2019open}
Milan Cvitkovic, Badal Singh, and Animashree Anandkumar.
\newblock Open vocabulary learning on source code with a graph-structured
  cache.
\newblock In \emph{International Conference on Machine Learning}, pages
  1475--1485. PMLR, 2019.

\bibitem[Dai et~al.(2019)Dai, Yang, Yang, Carbonell, Le, and
  Salakhutdinov]{dai2019transformer}
Zihang Dai, Zhilin Yang, Yiming Yang, Jaime Carbonell, Quoc~V Le, and Ruslan
  Salakhutdinov.
\newblock Transformer-xl: Attentive language models beyond a fixed-length
  context.
\newblock \emph{arXiv preprint arXiv:1901.02860}, 2019.

\bibitem[Devlin et~al.(2018)Devlin, Chang, Lee, and Toutanova]{devlin2018bert}
Jacob Devlin, Ming-Wei Chang, Kenton Lee, and Kristina Toutanova.
\newblock Bert: Pre-training of deep bidirectional transformers for language
  understanding.
\newblock \emph{arXiv preprint arXiv:1810.04805}, 2018.

\bibitem[Dinella et~al.(2020)Dinella, Dai, Li, Naik, Song, and
  Wang]{dinella2020hoppity}
Elizabeth Dinella, Hanjun Dai, Ziyang Li, Mayur Naik, Le~Song, and Ke~Wang.
\newblock Hoppity: Learning graph transformations to detect and fix bugs in
  programs.
\newblock In \emph{International Conference on Learning Representations
  (ICLR)}, 2020.

\bibitem[Ehrlinger and W{\"o}{\ss}(2016)]{ehrlinger2016towards}
Lisa Ehrlinger and Wolfram W{\"o}{\ss}.
\newblock Towards a definition of knowledge graphs.
\newblock \emph{SEMANTiCS (Posters, Demos, SuCCESS)}, 48:\penalty0 1--4, 2016.

\bibitem[Elnaggar et~al.(2021)Elnaggar, Ding, Jones, Gibbs, Feher, Angerer,
  Severini, Matthes, and Rost]{DBLP:journals/corr/abs-2104-02443}
Ahmed Elnaggar, Wei Ding, Llion Jones, Tom Gibbs, Tamas Feher, Christoph
  Angerer, Silvia Severini, Florian Matthes, and Burkhard Rost.
\newblock Codetrans: Towards cracking the language of silicone's code through
  self-supervised deep learning and high performance computing.
\newblock \emph{CoRR}, abs/2104.02443, 2021.
\newblock URL \url{https://arxiv.org/abs/2104.02443}.

\bibitem[Feng et~al.(2020)Feng, Guo, Tang, Duan, Feng, Gong, Shou, Qin, Liu,
  Jiang, and Zhou]{DBLP:journals/corr/abs-2002-08155}
Zhangyin Feng, Daya Guo, Duyu Tang, Nan Duan, Xiaocheng Feng, Ming Gong, Linjun
  Shou, Bing Qin, Ting Liu, Daxin Jiang, and Ming Zhou.
\newblock Codebert: {A} pre-trained model for programming and natural
  languages.
\newblock \emph{CoRR}, abs/2002.08155, 2020.
\newblock URL \url{https://arxiv.org/abs/2002.08155}.

\bibitem[Franceschi et~al.(2019)Franceschi, Niepert, Pontil, and
  He]{franceschi2019learning}
Luca Franceschi, Mathias Niepert, Massimiliano Pontil, and Xiao He.
\newblock Learning discrete structures for graph neural networks.
\newblock In \emph{International conference on machine learning}, pages
  1972--1982. PMLR, 2019.

\bibitem[Guo et~al.(2020)Guo, Ren, Lu, Feng, Tang, Liu, Zhou, Duan,
  Svyatkovskiy, Fu, Tufano, Deng, Clement, Drain, Sundaresan, Yin, Jiang, and
  Zhou]{DBLP:journals/corr/abs-2009-08366}
Daya Guo, Shuo Ren, Shuai Lu, Zhangyin Feng, Duyu Tang, Shujie Liu, Long Zhou,
  Nan Duan, Alexey Svyatkovskiy, Shengyu Fu, Michele Tufano, Shao~Kun Deng,
  Colin~B. Clement, Dawn Drain, Neel Sundaresan, Jian Yin, Daxin Jiang, and
  Ming Zhou.
\newblock Graphcodebert: Pre-training code representations with data flow.
\newblock \emph{CoRR}, abs/2009.08366, 2020.
\newblock URL \url{https://arxiv.org/abs/2009.08366}.

\bibitem[Gupta et~al.(2017)Gupta, Pal, Kanade, and Shevade]{gupta2017deepfix}
Rahul Gupta, Soham Pal, Aditya Kanade, and Shirish Shevade.
\newblock Deepfix: Fixing common c language errors by deep learning.
\newblock In \emph{Proceedings of the Thirty-First AAAI Conference on
  Artificial Intelligence}, pages 1345--1351, 2017.

\bibitem[Hellendoorn and Devanbu(2017)]{hellendoorn2017deep}
Vincent~J Hellendoorn and Premkumar Devanbu.
\newblock Are deep neural networks the best choice for modeling source code?
\newblock In \emph{Proceedings of the 2017 11th Joint Meeting on Foundations of
  Software Engineering}, pages 763--773, 2017.

\bibitem[Hellendoorn et~al.(2019)Hellendoorn, Sutton, Singh, Maniatis, and
  Bieber]{hellendoorn2019global}
Vincent~J Hellendoorn, Charles Sutton, Rishabh Singh, Petros Maniatis, and
  David Bieber.
\newblock Global relational models of source code.
\newblock In \emph{International Conference on Learning Representations}, 2019.

\bibitem[Hindle et~al.(2012)Hindle, Barr, Su, Gabel, and
  Devanbu]{hindle2012naturalness}
Abram Hindle, Earl~T Barr, Zhendong Su, Mark Gabel, and Premkumar Devanbu.
\newblock On the naturalness of software.
\newblock In \emph{2012 34th International Conference on Software Engineering
  (ICSE)}, pages 837--847. IEEE, 2012.

\bibitem[Hinton et~al.(2012)Hinton, Srivastava, and Swersky]{hinton2012neural}
Geoffrey Hinton, Nitsh Srivastava, and Kevin Swersky.
\newblock Neural networks for machine learning.
\newblock \emph{Coursera, video lectures}, 264\penalty0 (1), 2012.

\bibitem[Jang et~al.(2016)Jang, Gu, and Poole]{jang2016categorical}
Eric Jang, Shixiang Gu, and Ben Poole.
\newblock Categorical reparameterization with gumbel-softmax.
\newblock \emph{arXiv preprint arXiv:1611.01144}, 2016.

\bibitem[Johnson(2017)]{johnson2016learning}
Daniel~D Johnson.
\newblock Learning graphical state transitions.
\newblock In \emph{International Conference on Learning Representations}, 2017.

\bibitem[Johnson et~al.(2020)Johnson, Larochelle, and
  Tarlow]{johnson2020learning}
Daniel~D Johnson, Hugo Larochelle, and Daniel Tarlow.
\newblock Learning graph structure with a finite-state automaton layer.
\newblock \emph{arXiv preprint arXiv:2007.04929}, 2020.

\bibitem[Kanade et~al.(2020)Kanade, Maniatis, Balakrishnan, and
  Shi]{kanade2020learning}
Aditya Kanade, Petros Maniatis, Gogul Balakrishnan, and Kensen Shi.
\newblock Learning and evaluating contextual embedding of source code.
\newblock In \emph{International Conference on Machine Learning}, pages
  5110--5121. PMLR, 2020.

\bibitem[Karampatsis and Sutton(2019)]{karampatsis2019maybe}
Rafael-Michael Karampatsis and Charles Sutton.
\newblock Maybe deep neural networks are the best choice for modeling source
  code.
\newblock \emph{arXiv preprint arXiv:1903.05734}, 2019.

\bibitem[Karampatsis et~al.(2020)Karampatsis, Babii, Robbes, Sutton, and
  Janes]{karampatsis2020big}
Rafael-Michael Karampatsis, Hlib Babii, Romain Robbes, Charles Sutton, and
  Andrea Janes.
\newblock Big code!= big vocabulary: Open-vocabulary models for source code.
\newblock \emph{arXiv preprint arXiv:2003.07914}, 2020.

\bibitem[Kim et~al.(2020)Kim, Zhao, Tian, and Chandra]{kim2020code}
Seohyun Kim, Jinman Zhao, Yuchi Tian, and Satish Chandra.
\newblock Code prediction by feeding trees to transformers.
\newblock \emph{arXiv preprint arXiv:2003.13848}, 2020.

\bibitem[Kingma and Ba(2014)]{kingma2014adam}
Diederik~P Kingma and Jimmy Ba.
\newblock Adam: A method for stochastic optimization.
\newblock \emph{arXiv preprint arXiv:1412.6980}, 2014.

\bibitem[Kipf et~al.(2018)Kipf, Fetaya, Wang, Welling, and
  Zemel]{kipf2018neural}
Thomas Kipf, Ethan Fetaya, Kuan-Chieh Wang, Max Welling, and Richard Zemel.
\newblock Neural relational inference for interacting systems.
\newblock \emph{arXiv preprint arXiv:1802.04687}, 2018.

\bibitem[Koh et~al.(2020)Koh, Sagawa, Marklund, Xie, Zhang, Balsubramani, Hu,
  Yasunaga, Phillips, Gao, et~al.]{koh2020wilds}
Pang~Wei Koh, Shiori Sagawa, Henrik Marklund, Sang~Michael Xie, Marvin Zhang,
  Akshay Balsubramani, Weihua Hu, Michihiro Yasunaga, Richard~Lanas Phillips,
  Irena Gao, et~al.
\newblock Wilds: A benchmark of in-the-wild distribution shifts.
\newblock \emph{arXiv preprint arXiv:2012.07421}, 2020.

\bibitem[Krizhevsky et~al.(2012)Krizhevsky, Sutskever, and
  Hinton]{krizhevsky2012imagenet}
Alex Krizhevsky, Ilya Sutskever, and Geoffrey~E Hinton.
\newblock Imagenet classification with deep convolutional neural networks.
\newblock \emph{Advances in neural information processing systems},
  25:\penalty0 1097--1105, 2012.

\bibitem[Lin et~al.(2017)Lin, Goyal, Girshick, He, and
  Doll{\'a}r]{lin2017focal}
Tsung-Yi Lin, Priya Goyal, Ross Girshick, Kaiming He, and Piotr Doll{\'a}r.
\newblock Focal loss for dense object detection.
\newblock In \emph{Proceedings of the IEEE international conference on computer
  vision}, pages 2980--2988, 2017.

\bibitem[Luong et~al.(2015)Luong, Pham, and Manning]{luong2015effective}
Minh-Thang Luong, Hieu Pham, and Christopher~D Manning.
\newblock Effective approaches to attention-based neural machine translation.
\newblock \emph{arXiv preprint arXiv:1508.04025}, 2015.

\bibitem[Maddison and Tarlow(2014)]{maddison2014structured}
Chris Maddison and Daniel Tarlow.
\newblock Structured generative models of natural source code.
\newblock In \emph{International Conference on Machine Learning}, pages
  649--657. PMLR, 2014.

\bibitem[Maddison et~al.(2016)Maddison, Mnih, and Teh]{maddison2016concrete}
Chris~J Maddison, Andriy Mnih, and Yee~Whye Teh.
\newblock The concrete distribution: A continuous relaxation of discrete random
  variables.
\newblock \emph{arXiv preprint arXiv:1611.00712}, 2016.

\bibitem[Moonen(2001)]{moonen2001generating}
Leon Moonen.
\newblock Generating robust parsers using island grammars.
\newblock In \emph{Proceedings Eighth Working Conference on Reverse
  Engineering}, pages 13--22. IEEE, 2001.

\bibitem[Moonen(2002)]{moonen2002lightweight}
Leon Moonen.
\newblock Lightweight impact analysis using island grammars.
\newblock In \emph{Proceedings 10th International Workshop on Program
  Comprehension}, pages 219--228. IEEE, 2002.

\bibitem[Mou et~al.(2016)Mou, Li, Zhang, Wang, and Jin]{mou2016convolutional}
Lili Mou, Ge~Li, Lu~Zhang, Tao Wang, and Zhi Jin.
\newblock Convolutional neural networks over tree structures for programming
  language processing.
\newblock In \emph{Proceedings of the AAAI Conference on Artificial
  Intelligence}, volume~30, 2016.

\bibitem[Nguyen and Nguyen(2015)]{nguyen2015graph}
Anh~Tuan Nguyen and Tien~N Nguyen.
\newblock Graph-based statistical language model for code.
\newblock In \emph{2015 IEEE/ACM 37th IEEE International Conference on Software
  Engineering}, volume~1, pages 858--868. IEEE, 2015.

\bibitem[Nielson et~al.(2004)Nielson, Nielson, and
  Hankin]{nielson2004principles}
Flemming Nielson, Hanne~R Nielson, and Chris Hankin.
\newblock \emph{Principles of program analysis}.
\newblock Springer Science \& Business Media, 2004.

\bibitem[Nilsson-Nyman et~al.(2008)Nilsson-Nyman, Ekman, and
  Hedin]{nilsson2008practical}
Emma Nilsson-Nyman, Torbj{\"o}rn Ekman, and G{\"o}rel Hedin.
\newblock Practical scope recovery using bridge parsing.
\newblock In \emph{International Conference on Software Language Engineering},
  pages 95--113. Springer, 2008.

\bibitem[Nivre et~al.(2016)Nivre, De~Marneffe, Ginter, Goldberg, Hajic,
  Manning, McDonald, Petrov, Pyysalo, Silveira, et~al.]{nivre2016universal}
Joakim Nivre, Marie-Catherine De~Marneffe, Filip Ginter, Yoav Goldberg, Jan
  Hajic, Christopher~D Manning, Ryan McDonald, Slav Petrov, Sampo Pyysalo,
  Natalia Silveira, et~al.
\newblock Universal dependencies v1: A multilingual treebank collection.
\newblock In \emph{Proceedings of the Tenth International Conference on
  Language Resources and Evaluation (LREC'16)}, pages 1659--1666, 2016.

\bibitem[Nivre et~al.(2020)Nivre, de~Marneffe, Ginter, Haji{\v{c}}, Manning,
  Pyysalo, Schuster, Tyers, and Zeman]{nivre2020universal}
Joakim Nivre, Marie-Catherine de~Marneffe, Filip Ginter, Jan Haji{\v{c}},
  Christopher~D Manning, Sampo Pyysalo, Sebastian Schuster, Francis Tyers, and
  Daniel Zeman.
\newblock Universal dependencies v2: An evergrowing multilingual treebank
  collection.
\newblock \emph{arXiv preprint arXiv:2004.10643}, 2020.

\bibitem[Parisotto et~al.(2016)Parisotto, Mohamed, Singh, Li, Zhou, and
  Kohli]{parisotto2016neuro}
Emilio Parisotto, Abdel-rahman Mohamed, Rishabh Singh, Lihong Li, Dengyong
  Zhou, and Pushmeet Kohli.
\newblock Neuro-symbolic program synthesis.
\newblock \emph{arXiv preprint arXiv:1611.01855}, 2016.

\bibitem[Paulus et~al.(2020)Paulus, Choi, Tarlow, Krause, and
  Maddison]{paulus2020gradient}
Max~B Paulus, Dami Choi, Daniel Tarlow, Andreas Krause, and Chris~J Maddison.
\newblock Gradient estimation with stochastic softmax tricks.
\newblock \emph{arXiv preprint arXiv:2006.08063}, 2020.

\bibitem[Rabinovich et~al.(2017)Rabinovich, Stern, and
  Klein]{rabinovich2017abstract}
Maxim Rabinovich, Mitchell Stern, and Dan Klein.
\newblock Abstract syntax networks for code generation and semantic parsing.
\newblock \emph{arXiv preprint arXiv:1704.07535}, 2017.

\bibitem[Radford et~al.(2018)Radford, Narasimhan, Salimans, and
  Sutskever]{radford2018improving}
Alec Radford, Karthik Narasimhan, Tim Salimans, and Ilya Sutskever.
\newblock Improving language understanding by generative pre-training.
\newblock 2018.

\bibitem[Radford et~al.(2019)Radford, Wu, Child, Luan, Amodei, and
  Sutskever]{radford2019language}
Alec Radford, Jeffrey Wu, Rewon Child, David Luan, Dario Amodei, and Ilya
  Sutskever.
\newblock Language models are unsupervised multitask learners.
\newblock \emph{OpenAI blog}, 1\penalty0 (8):\penalty0 9, 2019.

\bibitem[Ray et~al.(2016)Ray, Hellendoorn, Godhane, Tu, Bacchelli, and
  Devanbu]{ray2016naturalness}
Baishakhi Ray, Vincent Hellendoorn, Saheel Godhane, Zhaopeng Tu, Alberto
  Bacchelli, and Premkumar Devanbu.
\newblock On the" naturalness" of buggy code.
\newblock In \emph{2016 IEEE/ACM 38th International Conference on Software
  Engineering (ICSE)}, pages 428--439. IEEE, 2016.

\bibitem[Raychev et~al.(2015)Raychev, Vechev, and
  Krause]{raychev2015predicting}
Veselin Raychev, Martin Vechev, and Andreas Krause.
\newblock Predicting program properties from" big code".
\newblock \emph{ACM SIGPLAN Notices}, 50\penalty0 (1):\penalty0 111--124, 2015.

\bibitem[Reps(1998)]{reps1998program}
Thomas Reps.
\newblock Program analysis via graph reachability.
\newblock \emph{Information and software technology}, 40\penalty0
  (11-12):\penalty0 701--726, 1998.

\bibitem[Sadowski et~al.(2015)Sadowski, Van~Gogh, Jaspan, Soderberg, and
  Winter]{sadowski2015tricorder}
Caitlin Sadowski, Jeffrey Van~Gogh, Ciera Jaspan, Emma Soderberg, and Collin
  Winter.
\newblock Tricorder: Building a program analysis ecosystem.
\newblock In \emph{2015 IEEE/ACM 37th IEEE International Conference on Software
  Engineering}, volume~1, pages 598--608. IEEE, 2015.

\bibitem[Sennrich et~al.(2015)Sennrich, Haddow, and Birch]{sennrich2015neural}
Rico Sennrich, Barry Haddow, and Alexandra Birch.
\newblock Neural machine translation of rare words with subword units.
\newblock \emph{arXiv preprint arXiv:1508.07909}, 2015.

\bibitem[Shaw et~al.(2018)Shaw, Uszkoreit, and Vaswani]{shaw2018self}
Peter Shaw, Jakob Uszkoreit, and Ashish Vaswani.
\newblock Self-attention with relative position representations.
\newblock \emph{arXiv preprint arXiv:1803.02155}, 2018.

\bibitem[Shiv and Quirk(2019)]{shiv2019novel}
Vighnesh Shiv and Chris Quirk.
\newblock Novel positional encodings to enable tree-based transformers.
\newblock In \emph{Advances in Neural Information Processing Systems}, pages
  12081--12091, 2019.

\bibitem[Svyatkovskiy et~al.(2020)Svyatkovskiy, Deng, Fu, and
  Sundaresan]{svyatkovskiy2020intellicode}
Alexey Svyatkovskiy, Shao~Kun Deng, Shengyu Fu, and Neel Sundaresan.
\newblock Intellicode compose: Code generation using transformer.
\newblock In \emph{Proceedings of the 28th ACM Joint Meeting on European
  Software Engineering Conference and Symposium on the Foundations of Software
  Engineering}, pages 1433--1443, 2020.

\bibitem[Svyatkovskoy et~al.(2020)Svyatkovskoy, Lee, Hadjitofi, Riechert,
  Franco, and Allamanis]{svyatkovskoy2020fast}
Alexey Svyatkovskoy, Sebastian Lee, Anna Hadjitofi, Maik Riechert, Juliana
  Franco, and Miltiadis Allamanis.
\newblock Fast and memory-efficient neural code completion.
\newblock \emph{arXiv preprint arXiv:2004.13651}, 2020.

\bibitem[Tarlow et~al.(2020)Tarlow, Moitra, Rice, Chen, Manzagol, Sutton, and
  Aftandilian]{tarlow2020learning}
Daniel Tarlow, Subhodeep Moitra, Andrew Rice, Zimin Chen, Pierre-Antoine
  Manzagol, Charles Sutton, and Edward Aftandilian.
\newblock Learning to fix build errors with graph2diff neural networks.
\newblock In \emph{Proceedings of the IEEE/ACM 42nd International Conference on
  Software Engineering Workshops}, pages 19--20, 2020.

\bibitem[Van~Deursen and Kuipers(1999)]{van1999building}
Arie Van~Deursen and Tobias Kuipers.
\newblock Building documentation generators.
\newblock In \emph{Proceedings IEEE International Conference on Software
  Maintenance-1999 (ICSM'99).'Software Maintenance for Business Change'(Cat.
  No. 99CB36360)}, pages 40--49. IEEE, 1999.

\bibitem[Vaswani et~al.(2017)Vaswani, Shazeer, Parmar, Uszkoreit, Jones, Gomez,
  Kaiser, and Polosukhin]{vaswani2017attention}
Ashish Vaswani, Noam Shazeer, Niki Parmar, Jakob Uszkoreit, Llion Jones,
  Aidan~N Gomez, {\L}ukasz Kaiser, and Illia Polosukhin.
\newblock Attention is all you need.
\newblock In \emph{Advances in neural information processing systems}, pages
  5998--6008, 2017.

\bibitem[Velickovic et~al.(2020)Velickovic, Buesing, Overlan, Pascanu, Vinyals,
  and Blundell]{velickovic2020pointer}
Petar Velickovic, Lars Buesing, Matthew~C Overlan, Razvan Pascanu, Oriol
  Vinyals, and Charles Blundell.
\newblock Pointer graph networks.
\newblock \emph{stat}, 1050:\penalty0 11, 2020.

\bibitem[Wang et~al.(2019)Wang, Shin, Liu, Polozov, and
  Richardson]{wang2019rat}
Bailin Wang, Richard Shin, Xiaodong Liu, Oleksandr Polozov, and Matthew
  Richardson.
\newblock Rat-sql: Relation-aware schema encoding and linking for text-to-sql
  parsers.
\newblock \emph{arXiv preprint arXiv:1911.04942}, 2019.

\bibitem[Wang et~al.(2020)Wang, Shen, Li, and Jin]{wang2020towards}
Wenhan Wang, Sijie Shen, Ge~Li, and Zhi Jin.
\newblock Towards full-line code completion with neural language models.
\newblock \emph{arXiv preprint arXiv:2009.08603}, 2020.

\bibitem[Weir(1992)]{weir1992efficiency}
Carl Weir.
\newblock Efficiency, robustness and accuracy in picky chart parsing.
\newblock In \emph{ACL}, 1992.

\bibitem[Williams(1992)]{williams1992simple}
Ronald~J Williams.
\newblock Simple statistical gradient-following algorithms for connectionist
  reinforcement learning.
\newblock \emph{Machine learning}, 8\penalty0 (3-4):\penalty0 229--256, 1992.

\bibitem[Xiong et~al.(2020)Xiong, Yang, He, Zheng, Zheng, Xing, Zhang, Lan,
  Wang, and Liu]{xiong2020layer}
Ruibin Xiong, Yunchang Yang, Di~He, Kai Zheng, Shuxin Zheng, Chen Xing,
  Huishuai Zhang, Yanyan Lan, Liwei Wang, and Tie-Yan Liu.
\newblock On layer normalization in the transformer architecture.
\newblock \emph{arXiv preprint arXiv:2002.04745}, 2020.

\bibitem[Yamaguchi(2015)]{yamaguchi2015pattern}
Fabian Yamaguchi.
\newblock \emph{Pattern-based vulnerability discovery}.
\newblock PhD thesis, Nieders{\"a}chsische Staats-und
  Universit{\"a}tsbibliothek G{\"o}ttingen, 2015.

\bibitem[Yang et~al.(2019)Yang, Dai, Yang, Carbonell, Salakhutdinov, and
  Le]{yang2019xlnet}
Zhilin Yang, Zihang Dai, Yiming Yang, Jaime Carbonell, Russ~R Salakhutdinov,
  and Quoc~V Le.
\newblock Xlnet: Generalized autoregressive pretraining for language
  understanding.
\newblock In \emph{Advances in neural information processing systems}, pages
  5753--5763, 2019.

\bibitem[Yasunaga and Liang(2020)]{yasunaga2020graph}
Michihiro Yasunaga and Percy Liang.
\newblock Graph-based, self-supervised program repair from diagnostic feedback.
\newblock \emph{arXiv preprint arXiv:2005.10636}, 2020.

\bibitem[Yin and Neubig(2017)]{yin2017syntactic}
Pengcheng Yin and Graham Neubig.
\newblock A syntactic neural model for general-purpose code generation.
\newblock \emph{arXiv preprint arXiv:1704.01696}, 2017.

\bibitem[Yin et~al.(2018)Yin, Zhou, He, and Neubig]{yin2018structvae}
Pengcheng Yin, Chunting Zhou, Junxian He, and Graham Neubig.
\newblock Structvae: Tree-structured latent variable models for semi-supervised
  semantic parsing.
\newblock \emph{arXiv preprint arXiv:1806.07832}, 2018.

\bibitem[Zhang et~al.(2019)Zhang, Wang, Zhang, Sun, Wang, and
  Liu]{zhang2019novel}
Jian Zhang, Xu~Wang, Hongyu Zhang, Hailong Sun, Kaixuan Wang, and Xudong Liu.
\newblock A novel neural source code representation based on abstract syntax
  tree.
\newblock In \emph{2019 IEEE/ACM 41st International Conference on Software
  Engineering (ICSE)}, pages 783--794. IEEE, 2019.

\bibitem[Z{\"u}gner et~al.(2021)Z{\"u}gner, Kirschstein, Catasta, Leskovec, and
  G{\"u}nnemann]{2021languageagnostic}
Daniel Z{\"u}gner, Tobias Kirschstein, Michele Catasta, Jure Leskovec, and
  Stephan G{\"u}nnemann.
\newblock Language-agnostic representation learning of source code from
  structure and context.
\newblock In \emph{International Conference on Learning Representations}, 2021.
\newblock URL \url{https://openreview.net/forum?id=Xh5eMZVONGF}.

\end{thebibliography}
\bibliographystyle{plainnat}

\appendix

\newpage

\appendix
\gdef\thesection{Appendix \Alph{section}}
\clearpage
\section{Edge Prediction Results for Variable Misuse}\label{app:edge_prediction_varmisuse}
\begin{figure*}[ht]
\begin{minipage}[t]{0.325\linewidth}
\centering
{\includegraphics[width=0.98\textwidth]{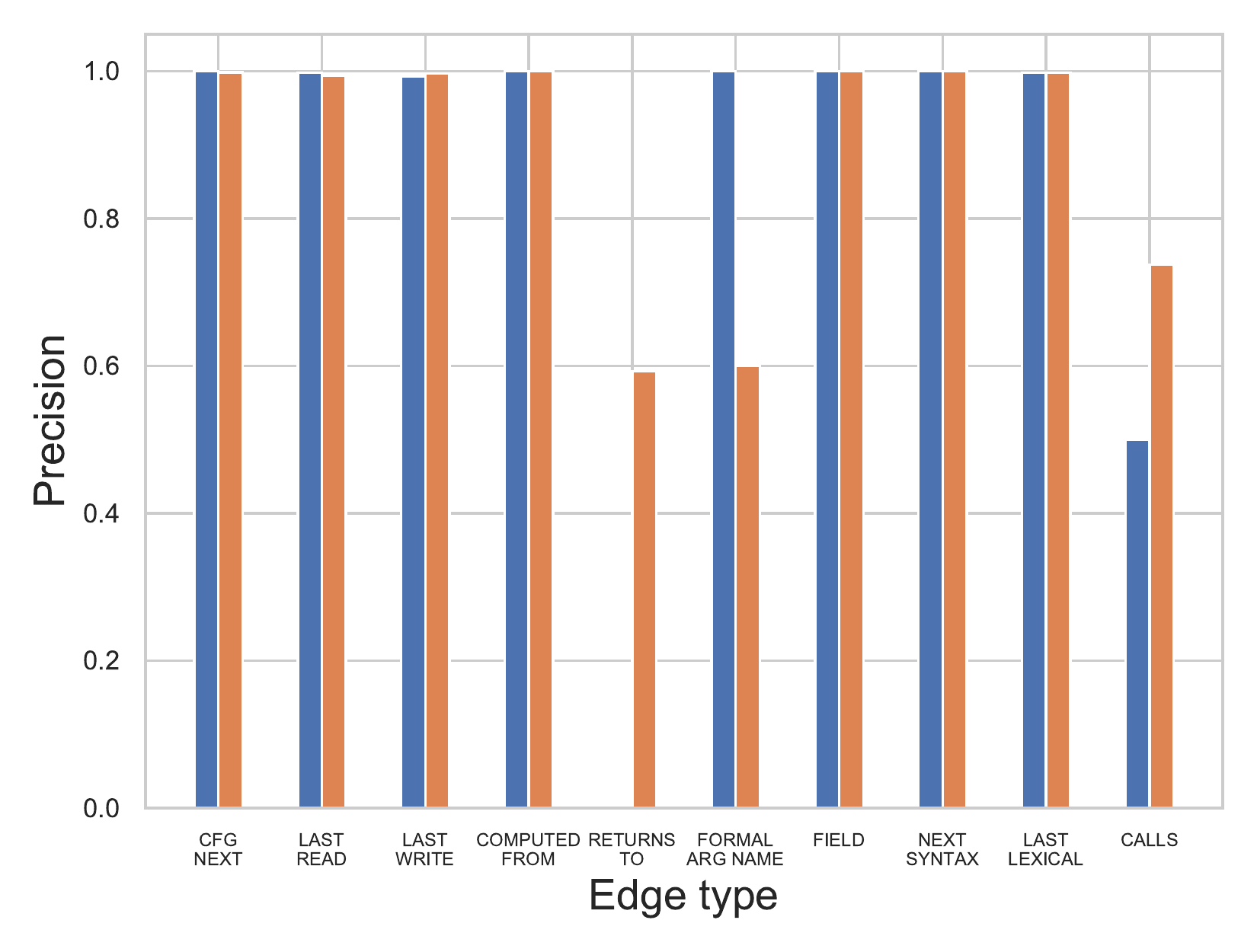}}
\\ \vspace{-0.10cm} (a) Precision
\end{minipage}
\begin{minipage}[t]{0.325\linewidth}
\centering
{\includegraphics[width=0.98\textwidth]{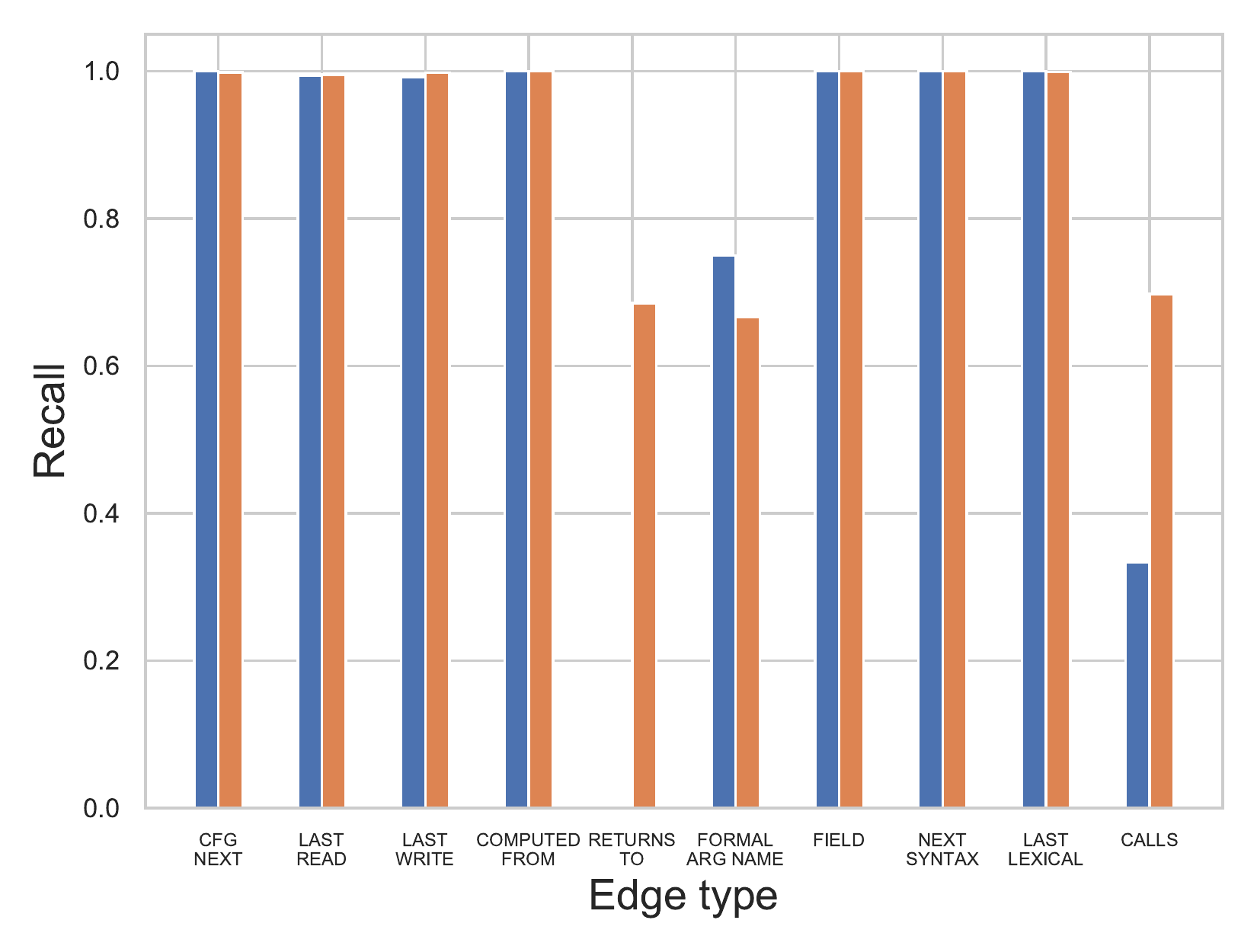}}
\\ \vspace{-0.10cm} (b) Recall
\end{minipage}
\begin{minipage}[t]{0.325\linewidth}
\centering
{\includegraphics[width=0.98\textwidth]{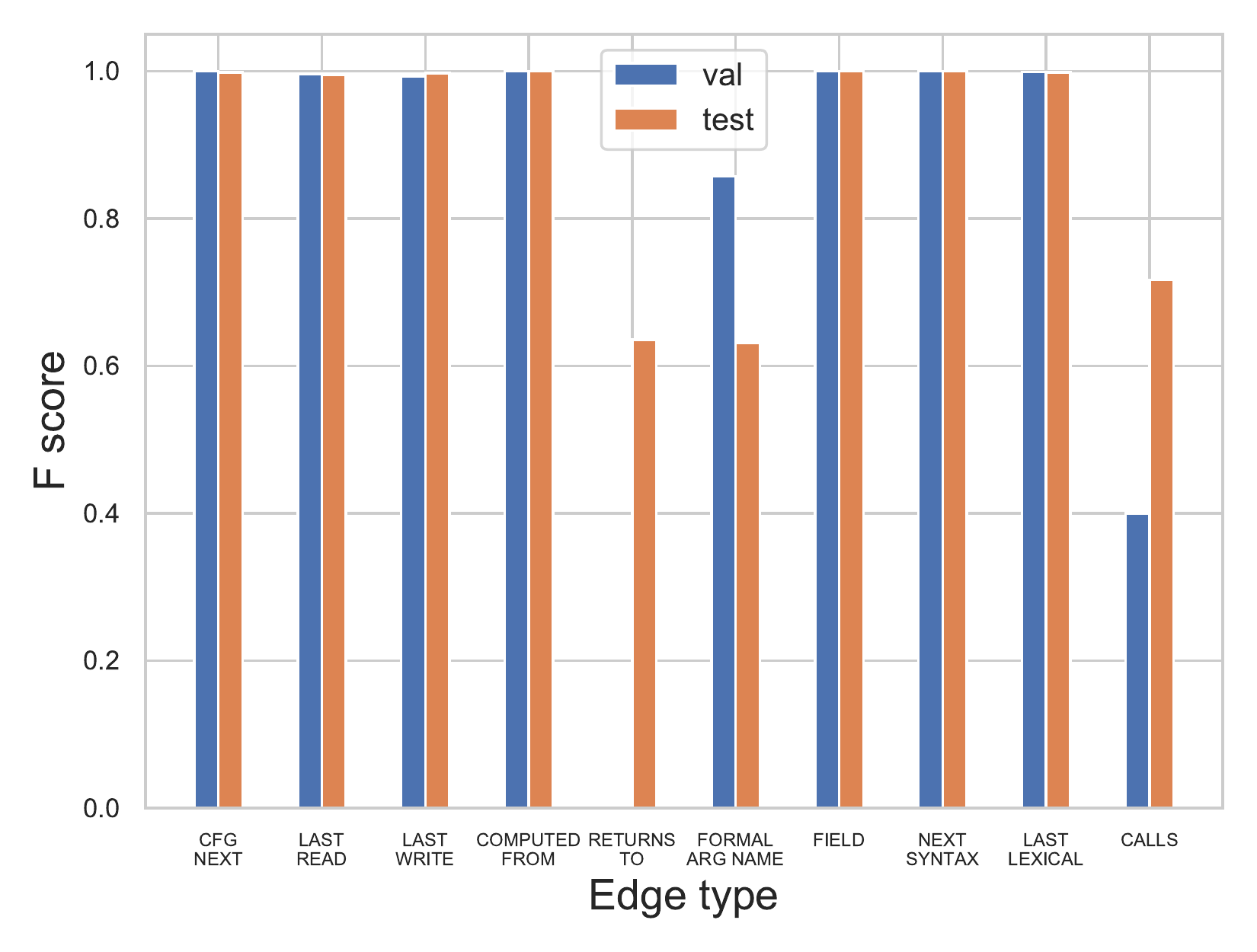}}
\\ \vspace{-0.10cm} (c) F score
\end{minipage}
\caption{
Results on the validation and test splits for the variable misuse dataset~\cite{hellendoorn2019global}. 
}
\label{fig:edge_prediction_varmisuse}
\end{figure*}
We include edge prediction results for clean code in the variable misuse task's dataset. 
In Figure~\ref{fig:edge_prediction_varmisuse}, we see compared to the results for the code completion task's dataset, the model performs slightly worse on various low frequency edge types. 

\clearpage
\section{Additional Results on Fine-tuning the Edge Prediction Model}\label{app:finetune}
\subsection{Setup for Fine-tuning the Edge Prediction Model}
For simplicity, given a well-trained edge prediction model, we fine-tune by jointly optimizing the loss with respect to parameters of the edge model and the task model with the same learning rate. 
We select the temperature $\tau$ based on validation performance, grid-searching over values in the range of $[0.01, 1]$. 

\subsection{Results on the Variable Misuse Task in the Literature}
Many papers in the literature report results for the variable misuse task based on clean examples. 
While our experiments in Section~\ref{subsec:finetune_edge} and the following also report standard metrics on this version, the results here only serve to improve our general understanding of relation model fine-tuning. 
Our primary focus is still on work-in-process code that may be broken or incomplete. 

We note that well-formed code is in general much easier to model, and it is therefore unsurprising that much better performance can be attained with additional tricks such as preprocessing code chunks into graphs after parsing them into ASTs~\cite{johnson2020learning} and more structured approaches designed with formal language theory in mind~\cite{johnson2020learning}. 
One may also expect that fine-tuning a large model pretrained on well-formed code to work well for tasks that also involve only well-formed code~\cite{kanade2020learning}. 

Lastly, we emphasize that our relation-aware model with ground truth edges results in accuracy numbers in the same range as those reported by~\citet{hellendoorn2019global} for clean code (our test accuracy is $72.7\%$ and $78.5\%$ for sequences with at most $512$ tokens, respectively for repair and localization; their reported results are $73.1\%$ and $76.9\%$ for sequences of fewer than $1000$ tokens).

\subsection{Plots Complementing Section~\ref{subsec:finetune_edge}}

\begin{figure*}[htpb]
\begin{minipage}[t]{0.48\linewidth}
\centering
{\includegraphics[width=0.7\textwidth]{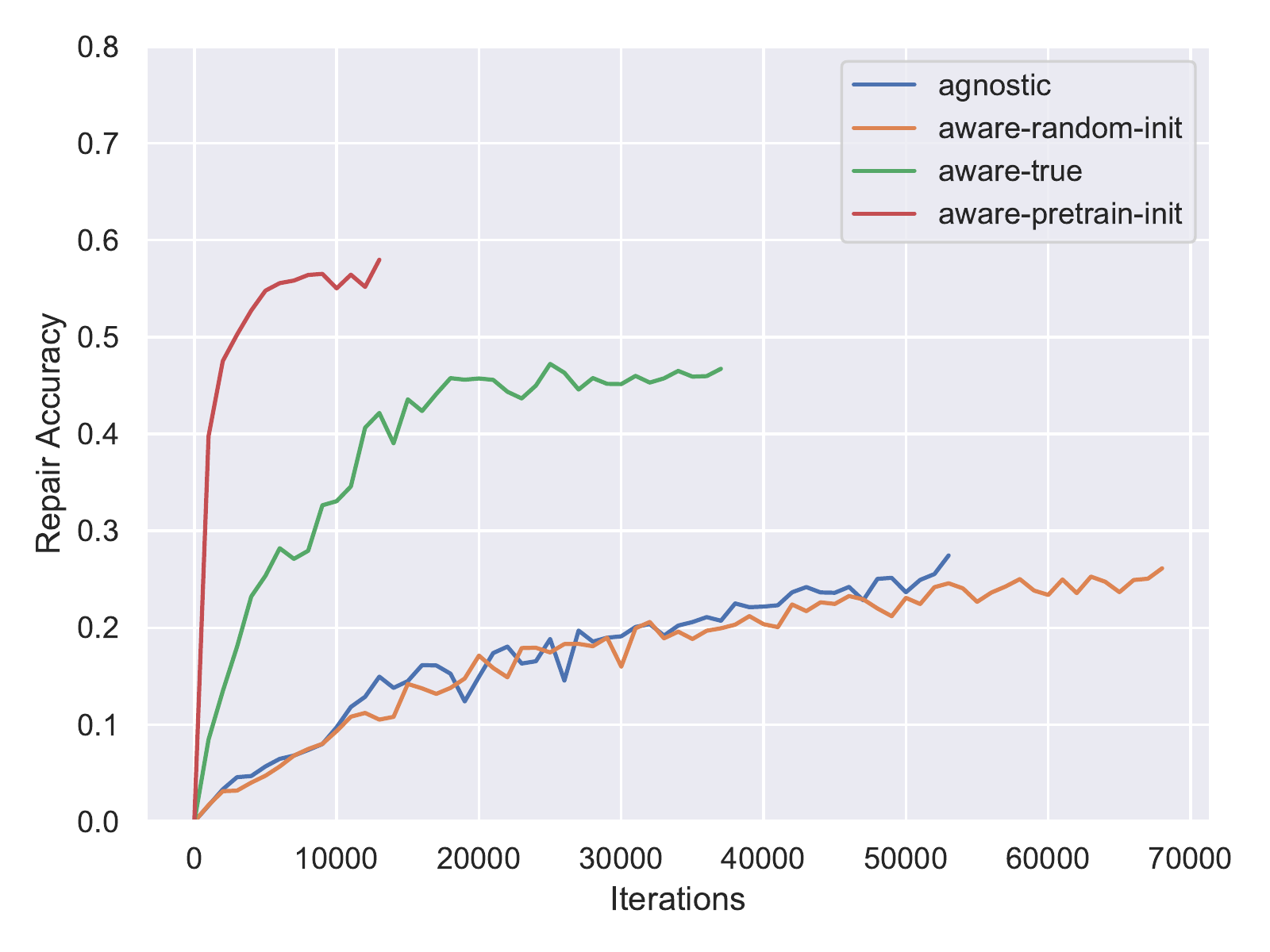}}
\\  (a) Repair Accuracy (Small Clean) 
\end{minipage}
\begin{minipage}[t]{0.48\linewidth}
\centering
{\includegraphics[width=0.7\textwidth]{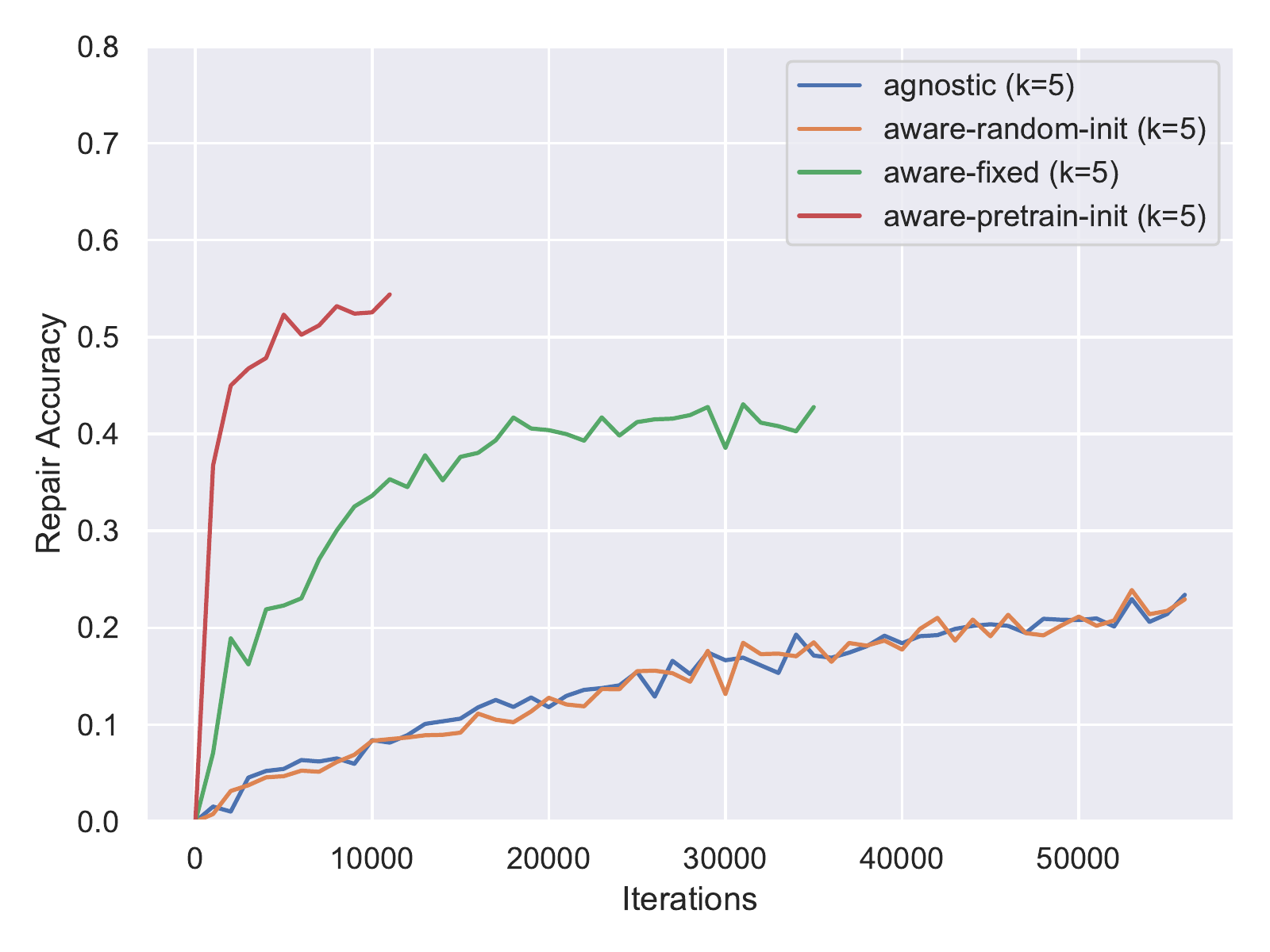}}
\\  (b) Repair Accuracy (Small WIP, $k=5$) 
\end{minipage}
\caption{
Test set repair accuracy for clean and work-in-progress code when edge model is fine-tuned on small training set. 
}
\label{fig:finetune_rep_small}
\end{figure*}

\begin{figure*}[htpb]
\begin{minipage}[t]{0.48\linewidth}
\centering
{\includegraphics[width=0.7\textwidth]{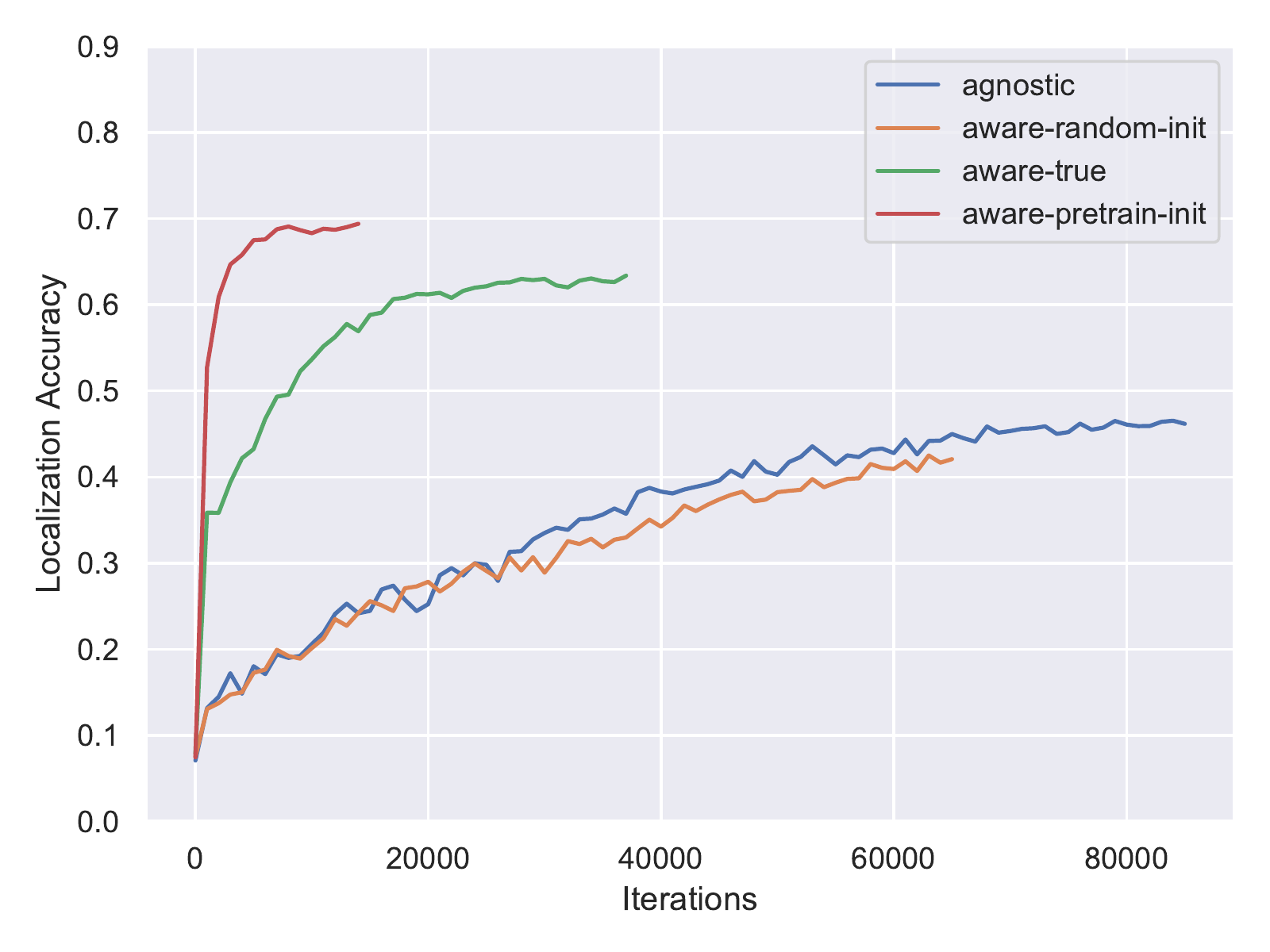}}
\\  (a) Localization Accuracy (Small Clean) 
\end{minipage}
\begin{minipage}[t]{0.48\linewidth}
\centering
{\includegraphics[width=0.7\textwidth]{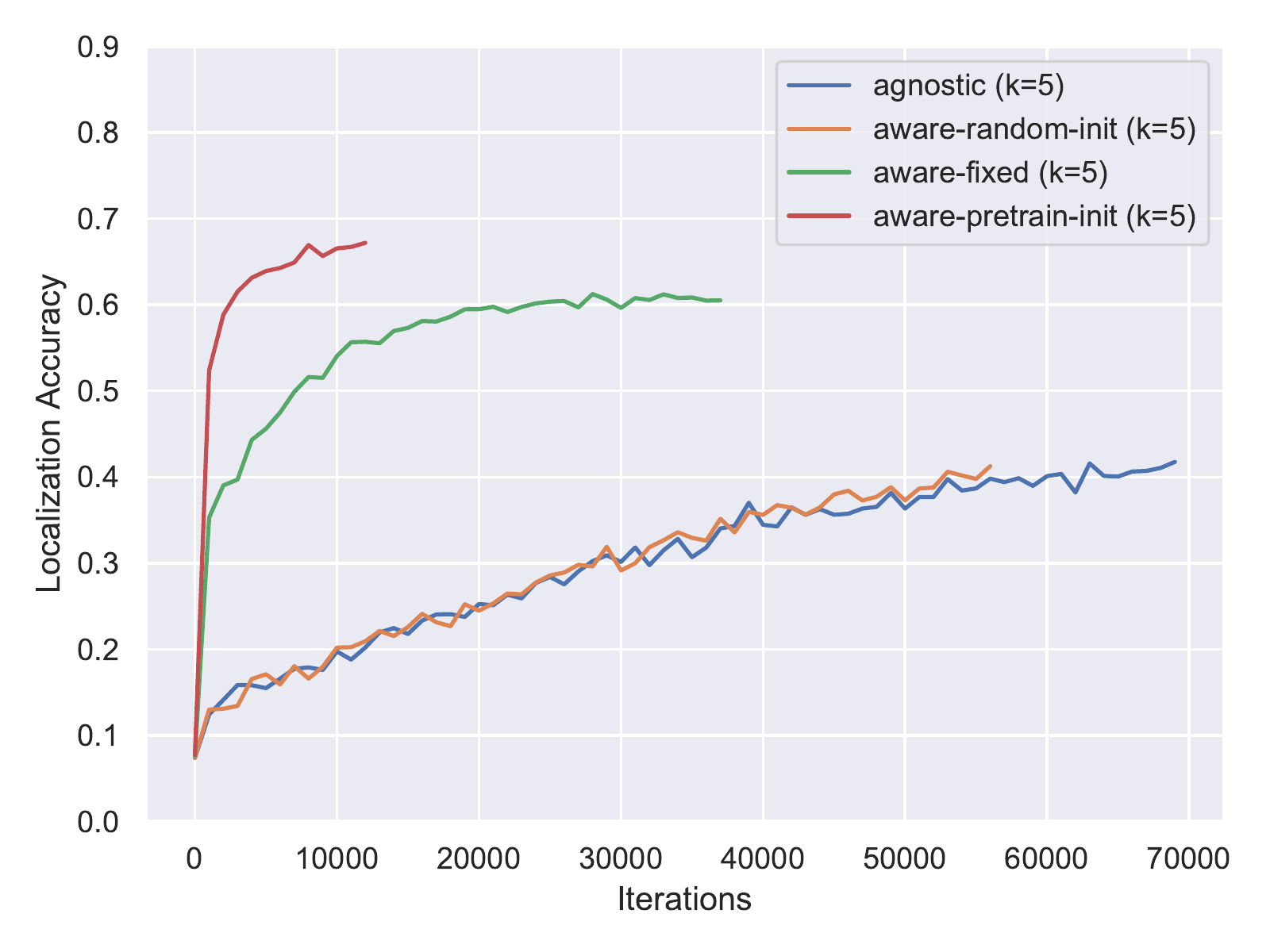}}
\\  (b) Localization Accuracy (Small WIP, $k=5$)
\end{minipage}
\caption{
Test set localization accuracy for clean and work-in-progress code when edge model is fine-tuned on small training set. 
}
\label{fig:finetune_loc_small}
\end{figure*}

\begin{figure*}[htpb]
\begin{minipage}[t]{0.48\linewidth}
\centering
{\includegraphics[width=0.7\textwidth]{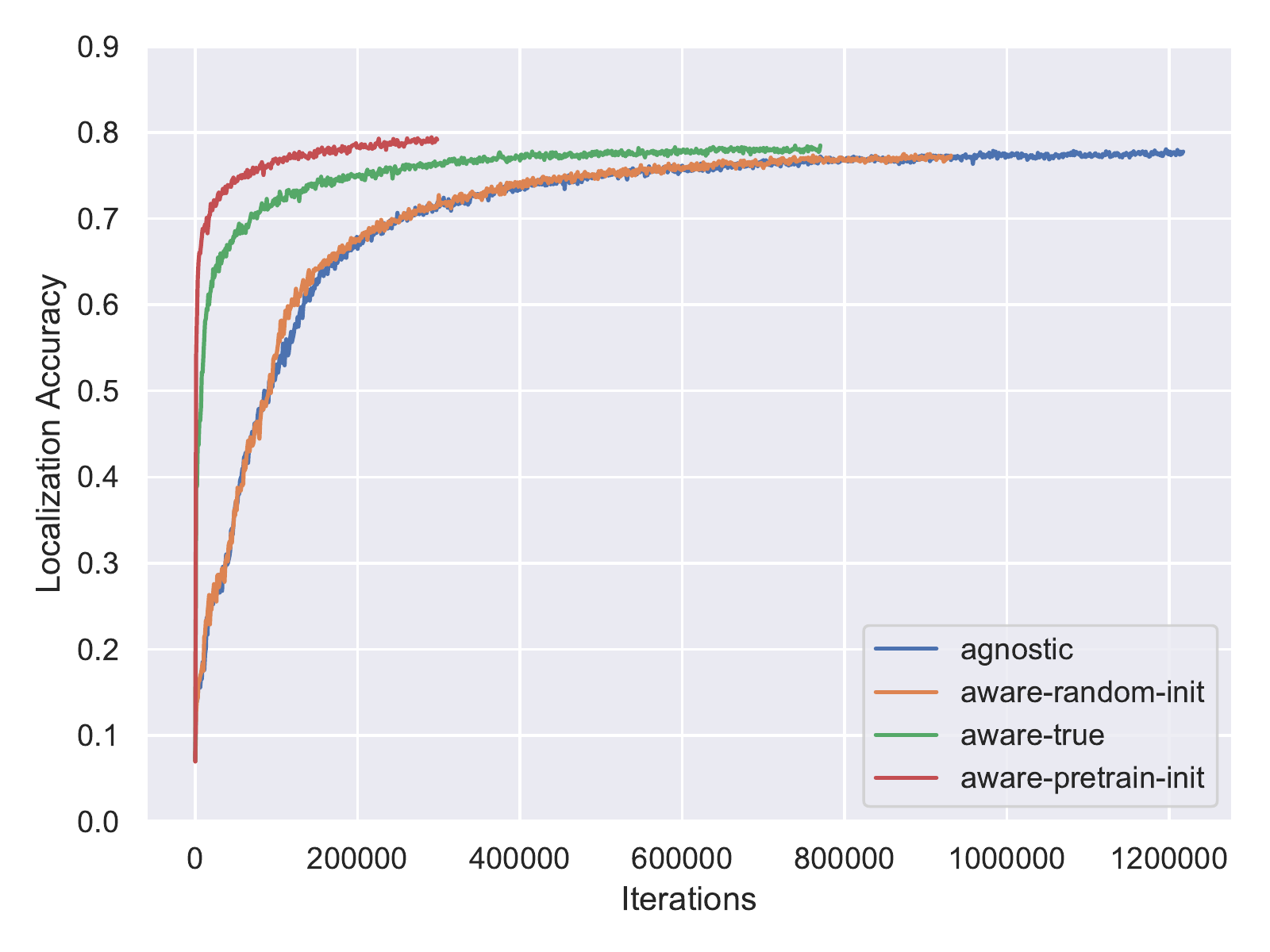}}
\\  (a) Localization Accuracy (Large Clean) 
\end{minipage}
\begin{minipage}[t]{0.48\linewidth}
\centering
{\includegraphics[width=0.7\textwidth]{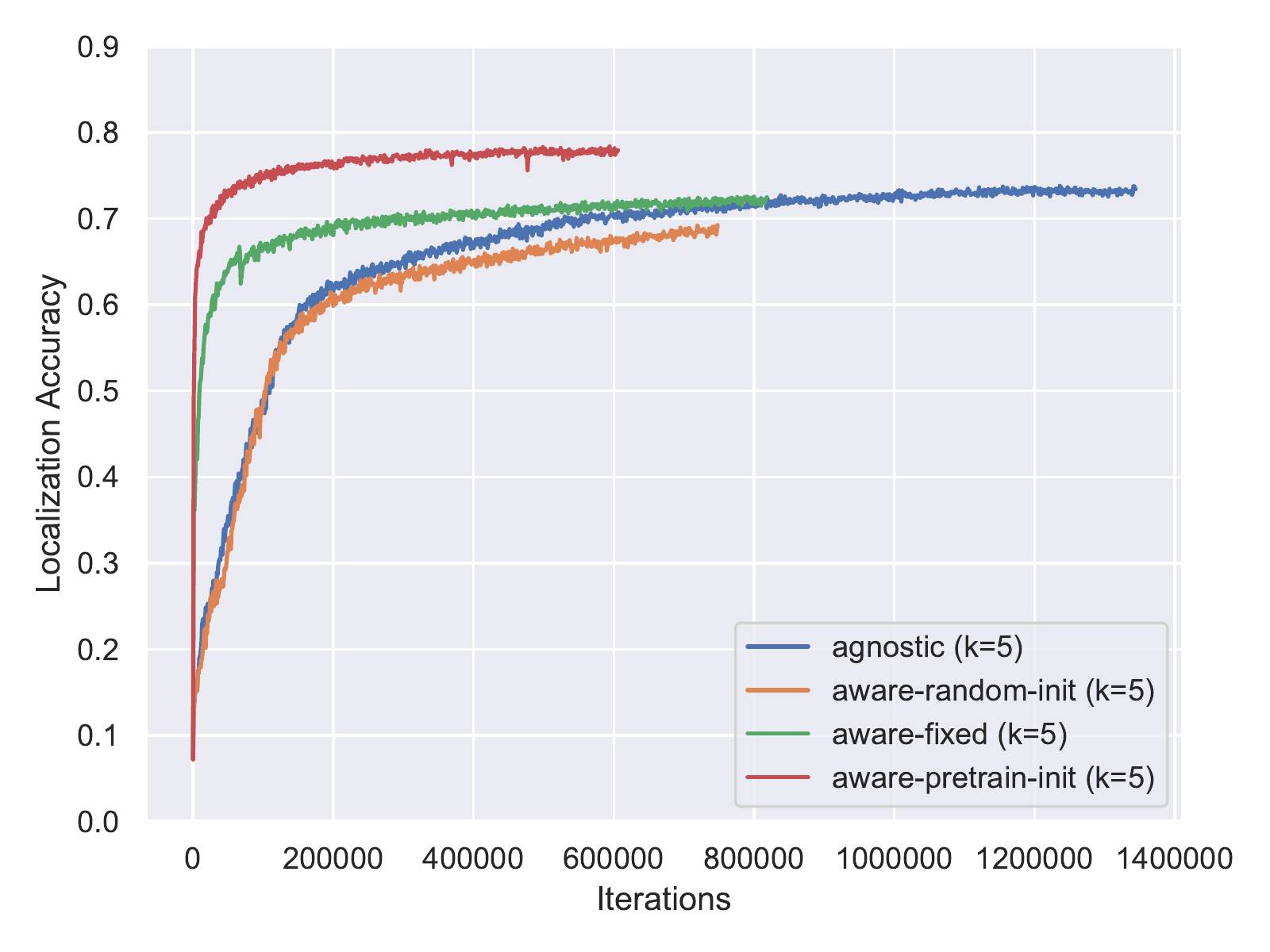}}
\\  (b) Localization Accuracy (Large WIP, $k=5$) 
\end{minipage}
\caption{
Test set localization accuracy for clean and work-in-progress code when edge model is fine-tuned on large training set. 
}
\label{fig:finetune_localization_large}
\end{figure*}

\newpage
\subsection{Additional Plots for WIP Code with $k=1$ and $k=2$}

\begin{figure*}[htpb]
\begin{minipage}[t]{0.48\linewidth}
\centering
{\includegraphics[width=0.7\textwidth]{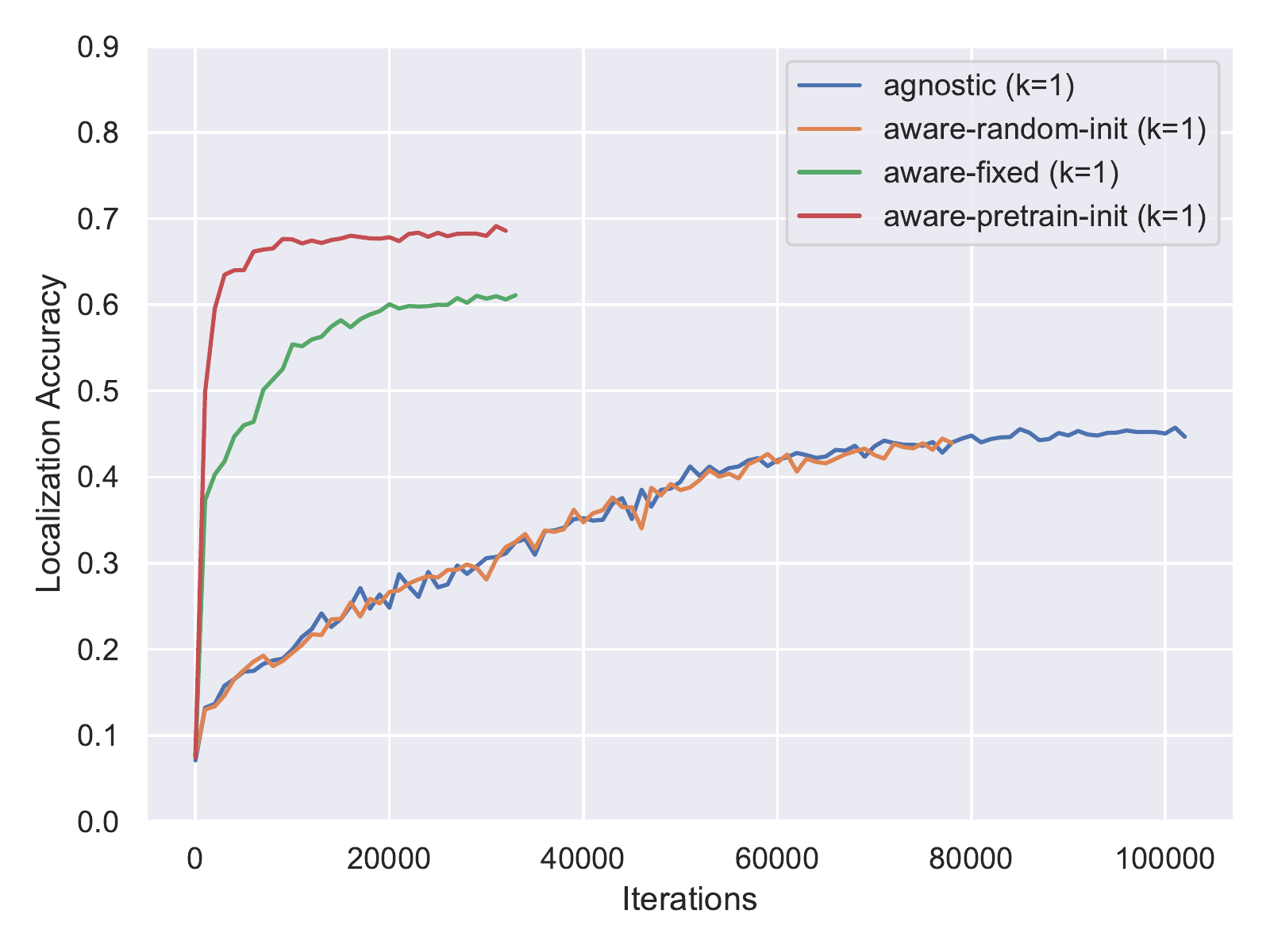}}
\\  (a) Localization Accuracy (Small WIP, $k=1$) 
\end{minipage}
\begin{minipage}[t]{0.48\linewidth}
\centering
{\includegraphics[width=0.7\textwidth]{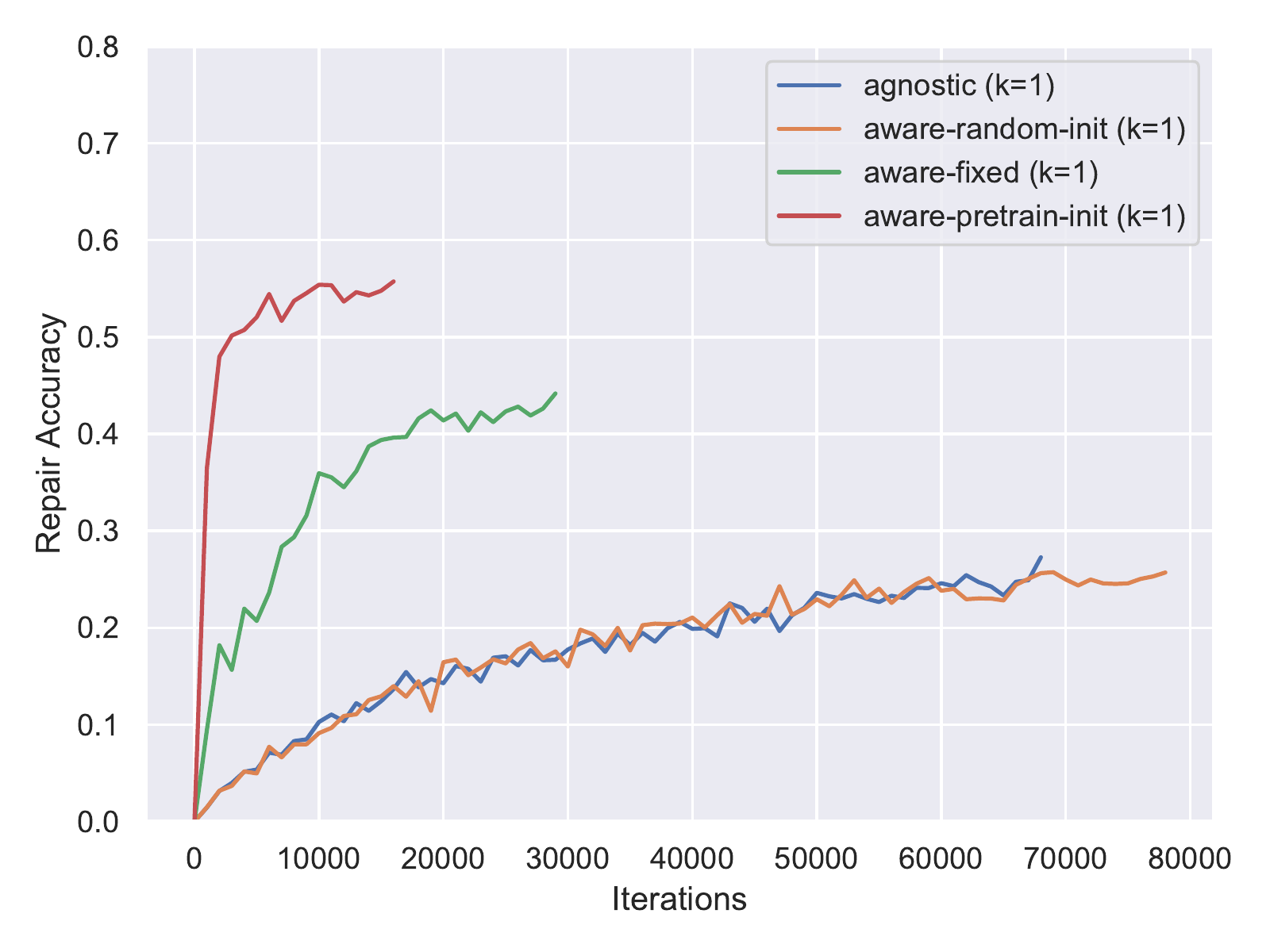}}
\\  (b) Repair Accuracy (Small WIP, $k=1$)
\end{minipage}

\begin{minipage}[t]{0.48\linewidth}
\centering
{\includegraphics[width=0.7\textwidth]{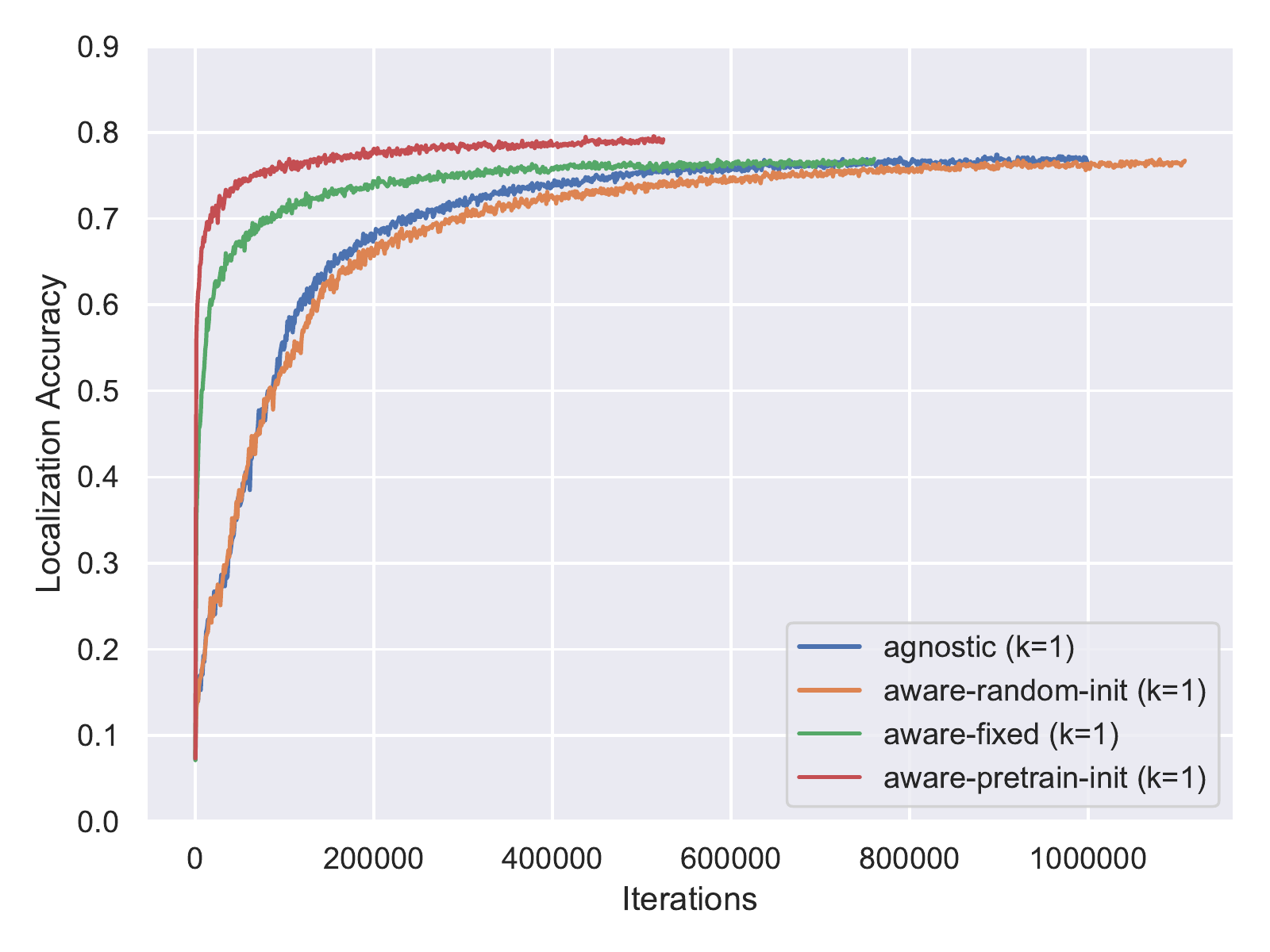}}
\\  (c) Localization Accuracy (Large WIP, $k=1$) 
\end{minipage}
\begin{minipage}[t]{0.48\linewidth}
\centering
{\includegraphics[width=0.7\textwidth]{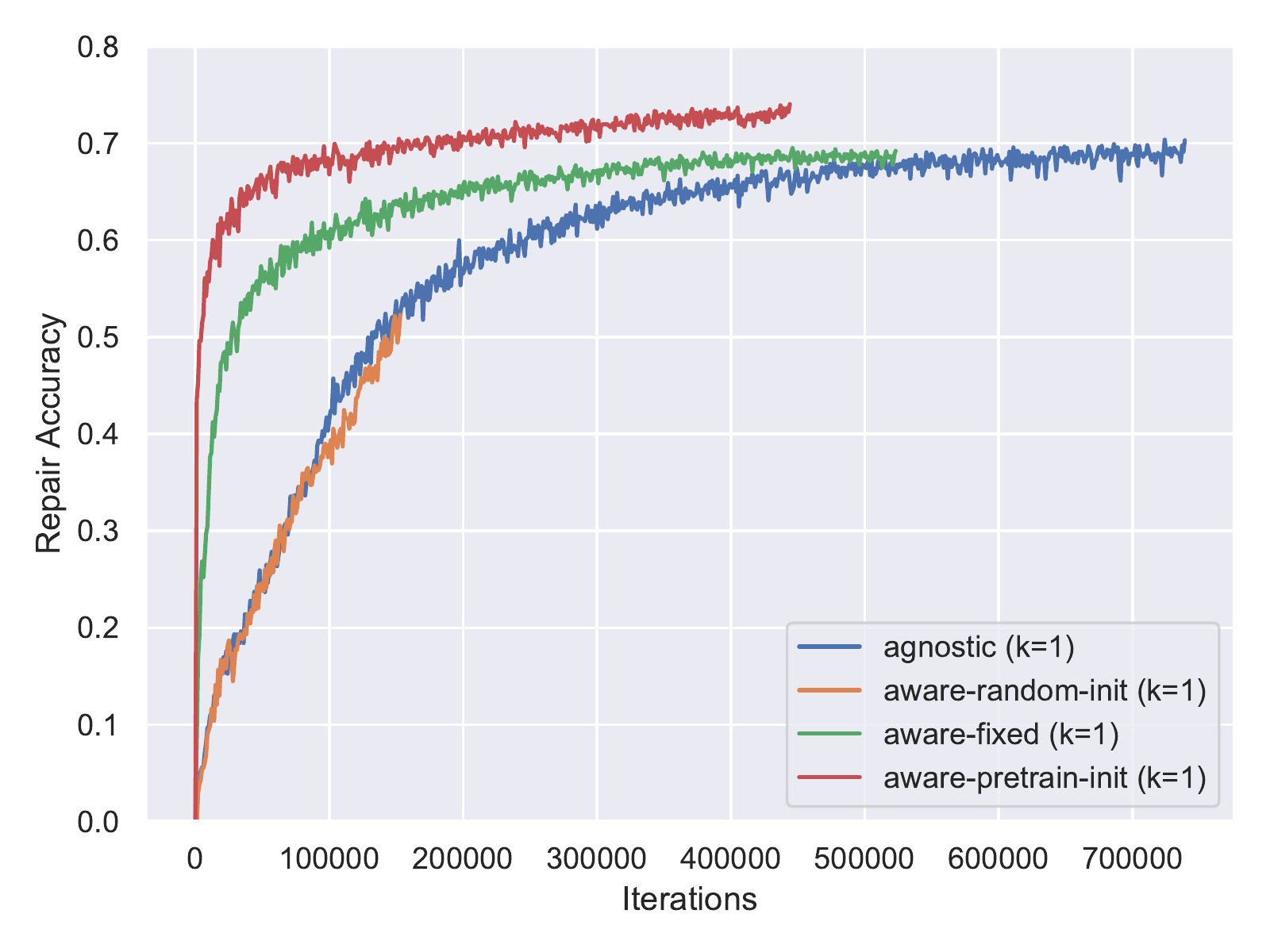}}
\\  (d) Repair Accuracy (Large WIP, $k=1$)
\end{minipage}
\caption{
Test set localization and repair accuracies for clean and work-in-progress code when edge model is fine-tuned.
}
\label{fig:finetune_rep_and_loc_small_k=1}
\end{figure*}

\begin{figure*}[htpb]
\begin{minipage}[t]{0.48\linewidth}
\centering
{\includegraphics[width=0.7\textwidth]{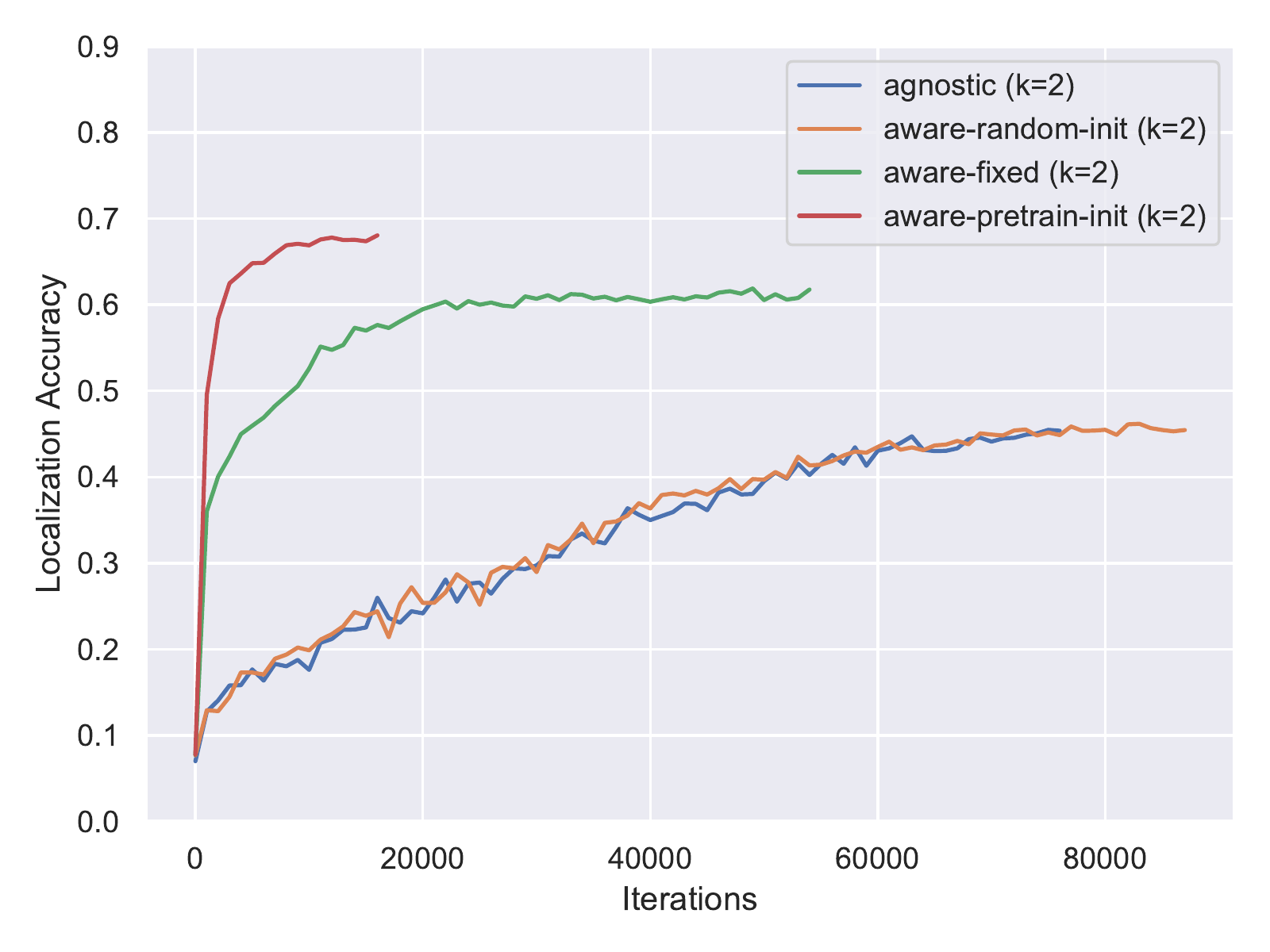}}
\\ (a) Localization Accuracy (Small WIP, $k=2$) 
\end{minipage}
\begin{minipage}[t]{0.48\linewidth}
\centering
{\includegraphics[width=0.7\textwidth]{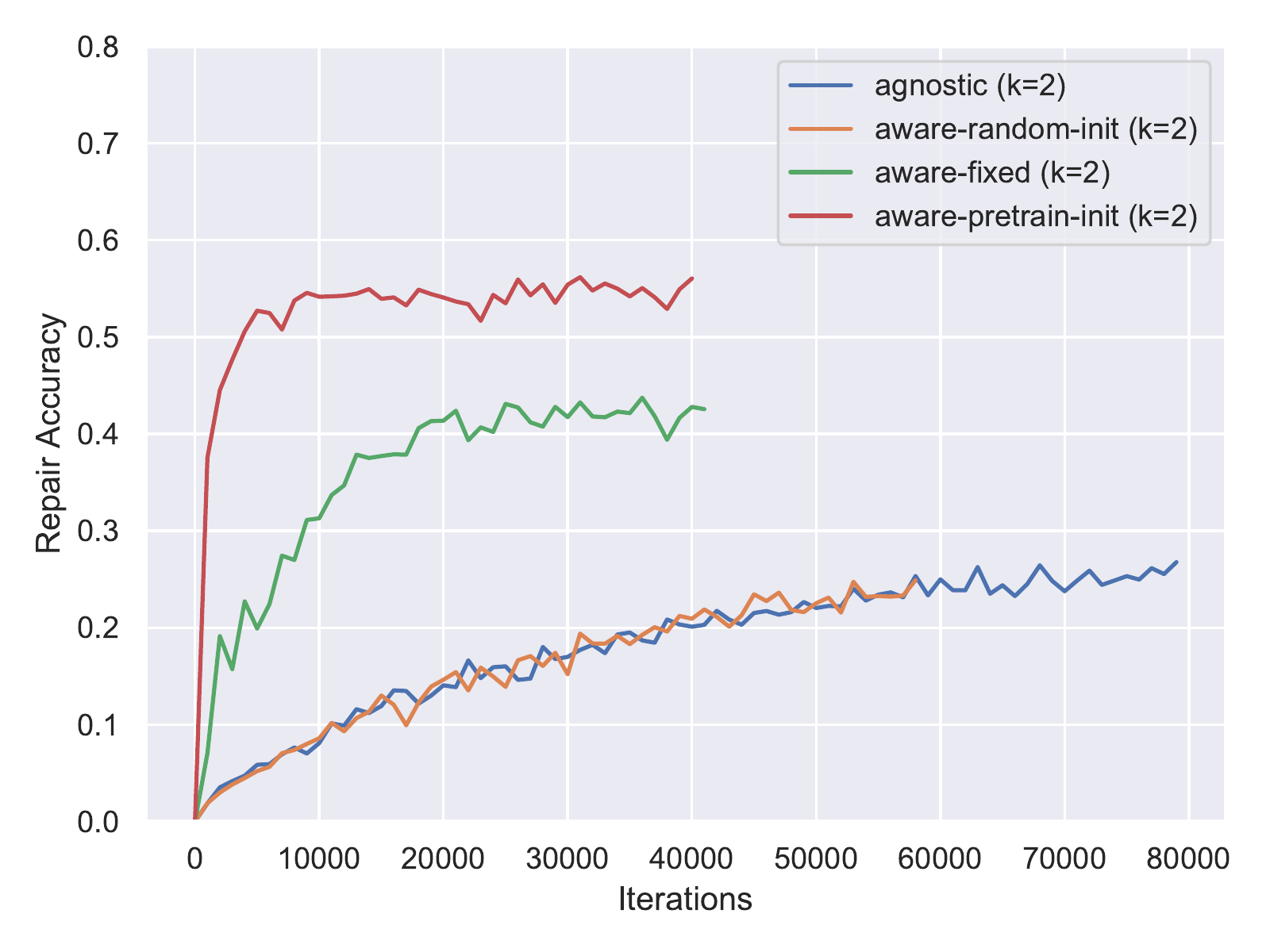}}
\\  (b) Repair Accuracy (Small WIP, $k=2$)
\end{minipage}

\begin{minipage}[t]{0.48\linewidth}
\centering
{\includegraphics[width=0.7\textwidth]{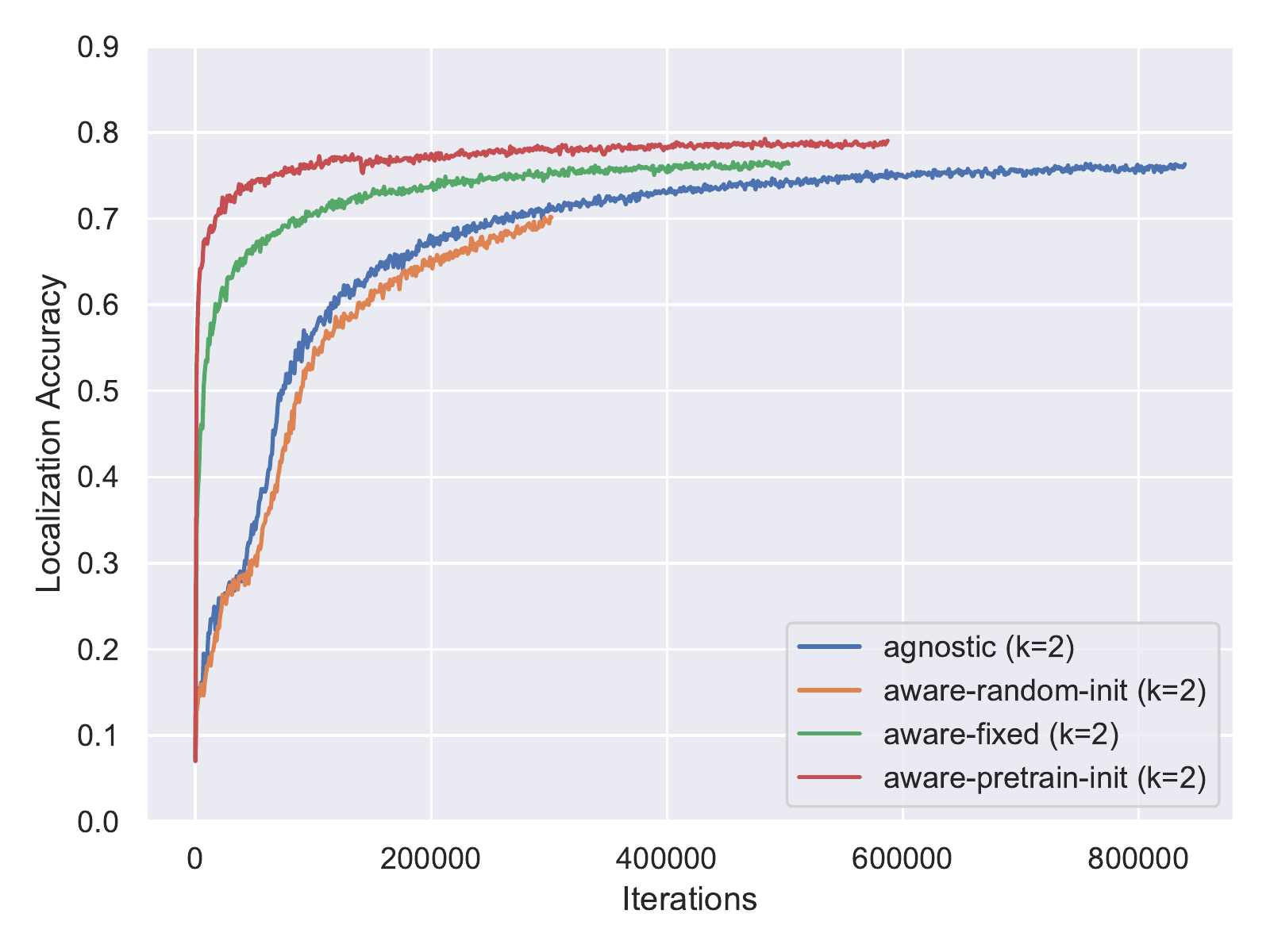}}
\\  (c) Localization Accuracy (Large WIP, $k=2$) 
\end{minipage}
\begin{minipage}[t]{0.48\linewidth}
\centering
{\includegraphics[width=0.7\textwidth]{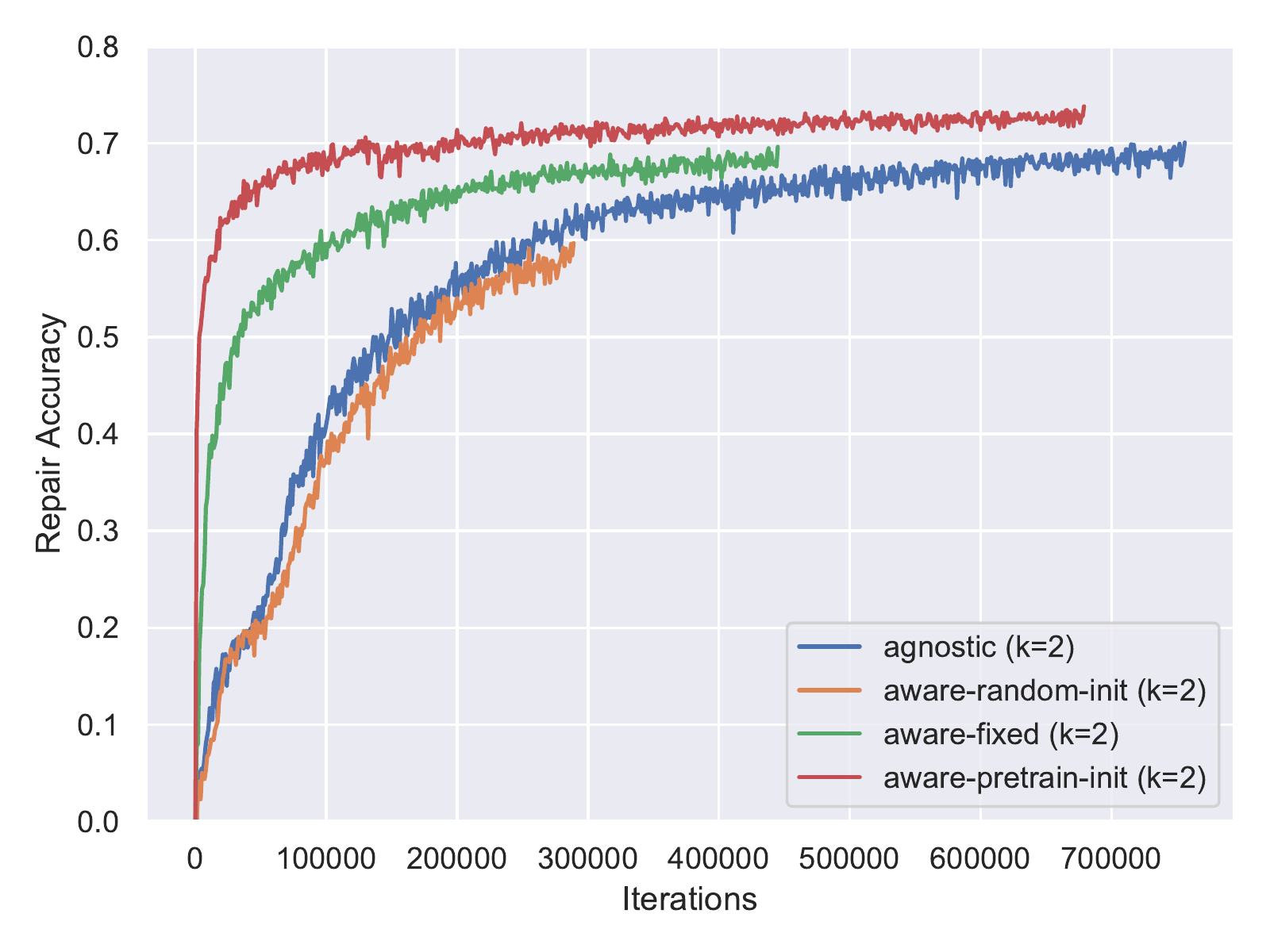}}
\\  (d) Repair Accuracy (Large WIP, $k=2$)
\end{minipage}
\caption{
Test set localization and repair accuracies for clean and work-in-progress code when edge model is fine-tuned.
}
\label{fig:finetune_rep_and_loc_small_k=2}
\end{figure*}

\clearpage
\section{Which Edges are Most Important for Code Completion? }\label{app:completion_ablation}

This section is devoted to studying which edge types are most helpful for the relation-aware model to achieve improved performance compared to its relation-agnostic counterpart on the code completion task. 
Meanwhile, we also uncover a reason behind the mediocre performance of relation-aware models when the fed edges are fixed and come from a learned edge prediction model. 

\begin{figure}[th]
\begin{minipage}[t]{0.49\linewidth}
\centering
{\includegraphics[width=0.8\textwidth]{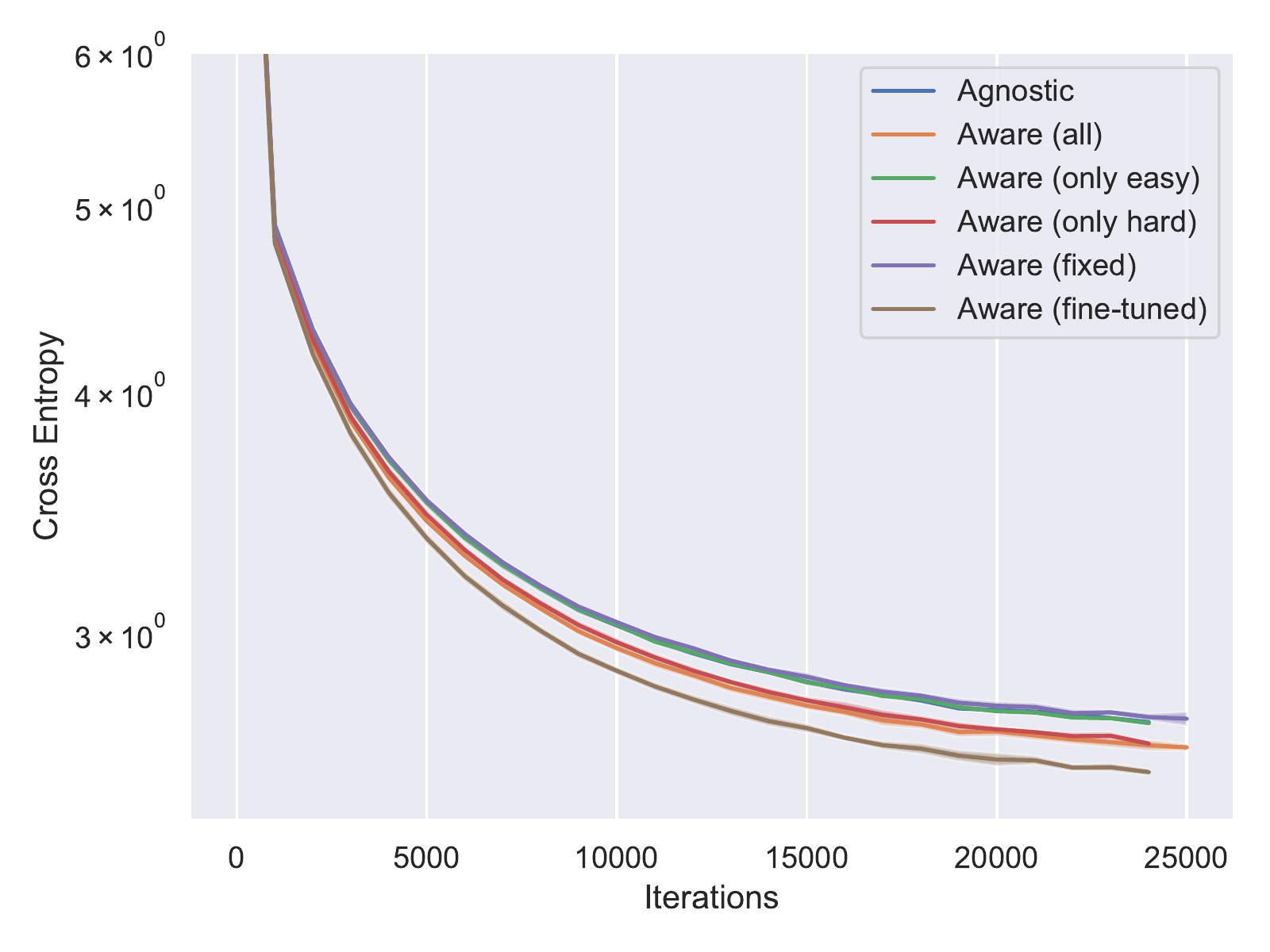}}
\\  (a) Cross entropy
\end{minipage}
\begin{minipage}[t]{0.49\linewidth}
\centering
{\includegraphics[width=0.8\textwidth]{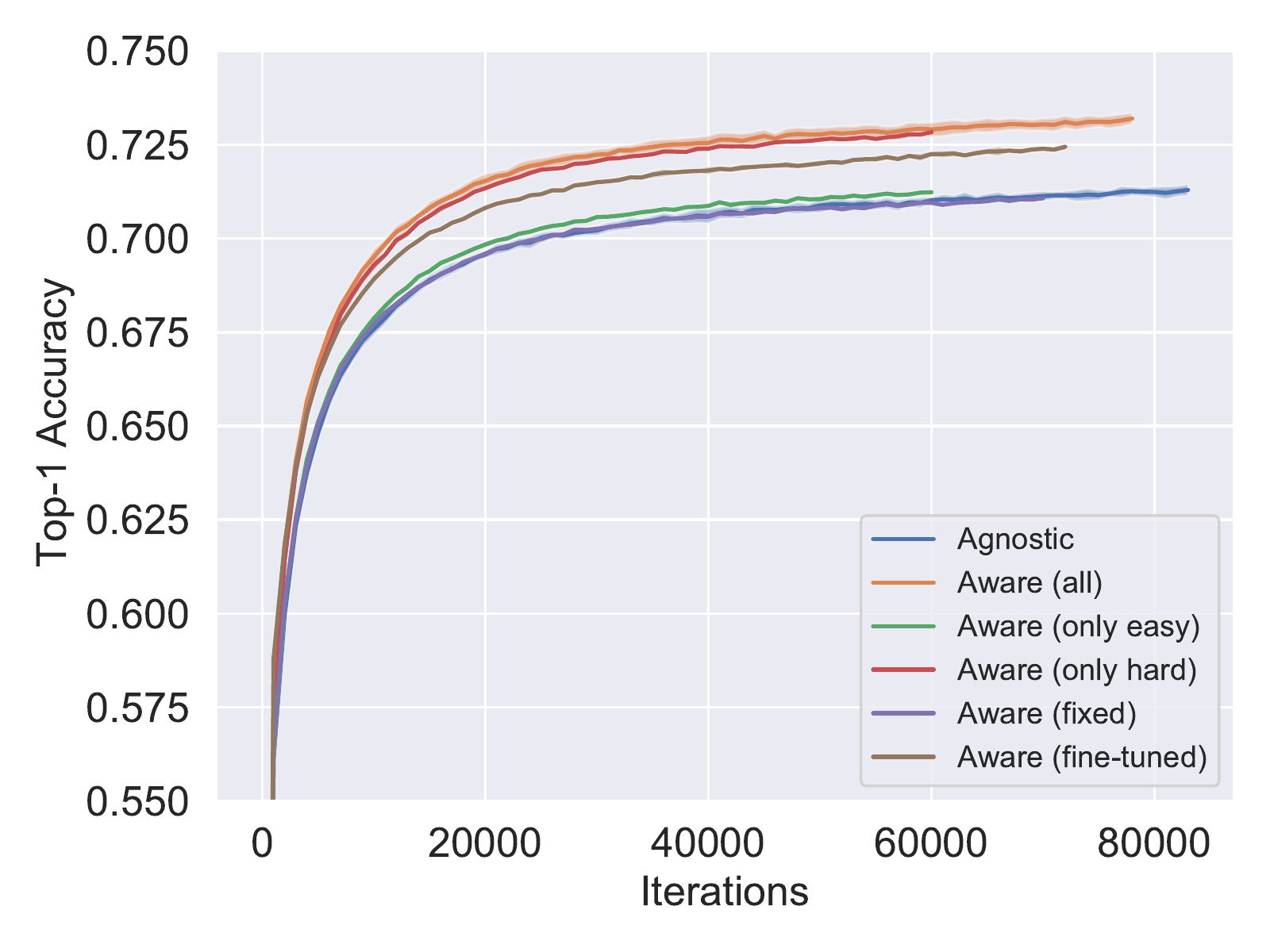}}
\\  (b) Top-1 accuracy
\end{minipage}
\caption{
Results on next token prediction for models trained on the small training split. 
``Aware (only easy)'' is nearly the same as ``Agnostic'', and ``Aware (only hard)'' is nearly the same as ``Aware (all)''. 
Results based on early stopping with a patience of 10k iterations on the validation split. 
}
\label{fig:code_completion_ablation}
\end{figure}

We split all edge types described in Section~\ref{subsec:ground_truth_relations} into those that are easy to predict (5 types
with highest validation F score) and those that are hard to predict (5 remaining types) based on the predictions of an edge prediction model trained as according to Section~\ref{subsec:learn_edges}. 
Then, we train relation-aware models using only ground truth edges from the former or latter category on the small training split. 
Figure~\ref{fig:code_completion_ablation} demonstrates that the downstream performance of the relation-aware model using only the easy-to-predict edges is nearly the same as the relation-agnostic model, and that using only hard-to-predict edges is nearly the same as the relation-aware model trained with all edge types. 

This experiment confirms the intuition that downstream accuracy is non-uniformly affected by different edge types and explains the poor performance of using our fixed edge prediction model. 

Lastly, we comment that the edge types most difficult to predict are CFGNext, ReturnsTo,
FormalArgName, Field, and Calls. 
These edges are derived from the associated control flow and data flow graphs of a piece of code.

\clearpage

\section{Subword-level Results for Code Completion}\label{app:completion_subword}
\begin{table*}[htbp]
\setlength\tabcolsep{1.6pt}
\centering
\begin{tabular}{c cccc}
\toprule
\multirow{2}[2]{*}{Model} & \multicolumn{2}{c}{\text{Perplexity ($\downarrow$)}}
                          & \multicolumn{2}{c}{\text{Top-1 accuracy in $\%$ ($\uparrow$)}} \\
\cmidrule(lr){2-3} \cmidrule(lr){4-5} & Small & Large & Small & Large \\
\midrule
{\footnotesize Transformer (stride=100) }   & $6.157 \pm 0.155$& $2.822 \pm 0.019$& $66.933 \pm 0.522$& $78.641 \pm 0.135$\\ 
{\footnotesize Transformer (stride=5) } & $\bm{6.128 \pm 0.155}$ & $2.806 \pm 0.019$& $66.973 \pm 0.529$& $78.741 \pm 0.137$\\ 
\midrule 
Agnostic   & $7.297 \pm 0.061$& $2.654 \pm 0.010$& $68.642 \pm 0.226$& $80.008 \pm 0.068$\\ 
Aware-fixed   & $7.217 \pm 0.016$& $2.663 \pm 0.006$& $69.053 \pm 0.097$& $79.922 \pm 0.027$\\ 
Aware-tuned   & $6.380 \pm 0.072$& $\bm{2.513 \pm 0.009} $& $\bm{69.740 \pm 0.099}$ & $\bm{81.035 \pm 0.072}$\\ 
\midrule 
Aware-true   & $6.690 \pm 0.083$& $2.535 \pm 0.002$& $70.226 \pm 0.155$& $81.223 \pm 0.012$ \\
\bottomrule
\end{tabular}
\caption{
Results for next subword prediction. 
Small and Large respectively refer to when the small and large training split is used. 
Numbers in bold are the best results in each column.
We exclude the Aware-true setting in our comparisons, since this setting cannot be executed in practice given any unseen code prefix. 
}
\label{table:completion_subword_level}
\end{table*}

\end{document}